\documentclass{article}
\usepackage{iclr2027_conference,times}
\usepackage{tabularx}
\usepackage{booktabs}
\usepackage{graphicx}
\usepackage{xcolor}
\usepackage{listings}
\usepackage{placeins}
\usepackage{wrapfig}
\lstdefinestyle{prompt}{
  basicstyle=\ttfamily\scriptsize,
  breaklines=true,
  columns=fullflexible,
  keepspaces=true,
  frame=single,
  framesep=4pt,
  xleftmargin=4pt,
  xrightmargin=4pt,
  showstringspaces=false,
}

\usepackage{amsmath,amsfonts,bm}

\def\eqref#1{equation~\ref{#1}}

\def\1{\bm{1}}

\DeclareMathAlphabet{\mathsfit}{\encodingdefault}{\sfdefault}{m}{sl}
\SetMathAlphabet{\mathsfit}{bold}{\encodingdefault}{\sfdefault}{bx}{n}

\usepackage{hyperref}
\usepackage{url}

\title{Model Internals Speak Volumes:\\J-Lens Loudness Predicts Decodability}

\author{Alexandru Dimofte$^1$, David Bau$^2$, Mario Giulianelli$^{3,4}$, Gabriele Sarti$^{2,3}$  \\
University of Zurich$^1$, Northeastern University$^2$, Parallax$^3$, University College London$^4$ \\
\\
\texttt{alexandru.dimofte@uzh.ch}\;\;
\texttt{d.bau@northeastern.edu}\\
\texttt{gsarti@parallax.ai}\;\; 
\texttt{m.giulianelli@ucl.ac.uk} 
}

\iclrfinalcopy
\begin{document}

\maketitle

\begin{abstract}
When probing the activations of a Large Language Model (LLM), several design decisions arise. Besides the probe design itself, one must choose which layer(s) and token position(s) to probe. These choices are often made heuristically or through expensive sweeps over candidate layers and positions, and to our knowledge, no method yet predicts the decodability of a signal directly from the activations. We take a step in this direction by introducing J-Lens loudness, a new application of the Jacobian lens (J-Lens) that guides these choices without labels or probe training. We represent a signal of interest by a small subset of the model's vocabulary, and define J-Lens loudness as the log probability mass that the J-Lens assigns to this subset at a given activation. In the reasoning chains of two LLM agents navigating grid worlds, the balanced accuracy of next-action probes increases steadily with direction signal loudness, and the same holds for a probe whose labels come from the environment rather than from the model. Loud positions are more decodable than random ones at every layer, and the loudest layer lies on the plateau of highest decodability found by a full probe sweep. Taken together, we show that J-Lens loudness can be used as a criterion to find token positions, and a suitable layer, whose activations contain underlying computation relevant to a signal of interest. We note that this is a work in progress and describe the limitations of this preprint in the the Discussion section under Limitations. Our code is available on \href{https://github.com/AlexDimofte19/Model-Internals-Speak-Volumes-J-Lens-Loudness-Predicts-Decodability}{GitHub}.

\end{abstract}

\section{Introduction}\label{sec:introduction}

In a time of rapid improvements in the performance of Large Language Models (LLMs), understanding the computations behind their behaviour has become increasingly important for the field of interpretability \citep{sharkey2025open}.

One of the main methods used across the field is probing \citep{alain2016understanding}. Researchers employing this method typically aim to test whether a signal, that is, information about some aspect of the input or the model's computation, such as a feature, or a property, is decodable from a model's activations at a specific layer and token position. Typically, one trains a mapping between those activations and labels for the signal of interest, ranging from a simple linear classifier to a more expressive model such as a Multilayer Perceptron (MLP).

A major practical difficulty of probing is that, before a probe can be trained, one must already know where the relevant signal is decodable, and a poor guess can make a signal look absent when it is not. The first choice is the layer. Conventional wisdom tells us that middle layers are typically where information is richest. Indeed, intermediate layers often yield more useful representations than final ones \citep{skean2025layer}, and probe performance frequently peaks there \citep{lugoloobi2026llms, gurnee2024language, marks2024geometry}, but where exactly the peak lies varies across models and signals.
The second choice is which token(s) to probe, and it is just as consequential. Information is distributed across tokens non-uniformly, and some signals concentrate at specific positions, such as exact-answer tokens for truthfulness \citep{orgad2025llms} or punctuation tokens for sentiment \citep{tigges2023linear}. Probe performance, therefore, depends heavily on the choice of token position, which is rarely known in advance.

These are important decisions which shape how we think about the inner workings of LLMs and are used in monitoring models, but they are often based on results which are not necessarily transferable between problems. The most common approach is to sweep over layers $L$ (and sometimes positions $P$), training a separate probe for every combination and keeping the best one. This is costly, requiring $O(L \times P)$ probes and labelled data for every signal. The alternative is to rely on heuristics such as just picking the middle layer, which is cheap but not signal-specific.

This calls for a new family of measurements of decodability which would tell us the degree to which a signal is encoded in a specific activation, without relying on probes. Luckily, such readouts of intermediate activations already exist in the form of Logit Lens \citep{nostalgebraist2020logitlens} and Jacobian Lens (J-Lens) \citep{gurnee2026verbalizable}, but to our knowledge, neither has yet been examined on their capabilities as signal-specific predictors of decodability across layers and token positions.

We introduce \emph{J-Lens loudness} to fill this gap. Our method relies primarily on the Jacobian Lens, which identifies the tokens a model is poised to verbalize at a given point in its processing \citep{gurnee2026verbalizable}. For a signal $s$ with an associated vocabulary subset $V_s$, we define J-Lens loudness as the log probability mass the J-Lens assigns to $V_s$ at the activation of a given layer and token position.

We compute loudness over multiple layer--position pairs in the reasoning chains of GPT-OSS-20B \citep{openai2025gptoss} and Qwen3.6-35B-A3B \citep{qwen36_35b_a3b} agents navigating 2D grid worlds \citep{arghal2026behavioural}. We mainly study a signal related to direction information, which we relate to computations about the next step. We evaluate a second signal related to the grid state which serves as a control. We choose the ``loudest'' layer for each signal and the ``loudest'' positions within each reasoning chain (Section~\ref{sec:methods}), and test whether the signal is more decodable there.

We make the following claims:
\begin{itemize}
    \item J-Lens loudness is a criterion that requires no probe training or labels and can be used to find the token positions from which a signal is decodable, and a layer on the plateau of highest decodability; the selected layer is robust to perturbations of the signal vocabulary;
    \item Decodability increases steadily with J-Lens loudness, even at positions where no signal words are written, both for labels derived from the model's output and for labels derived from the environment;
    \item J-Lens loudness is selective: the decodability of a signal $s$ is predicted by the loudness of $s$, but not by the loudness of an unrelated signal $s'$;
    \item J-Lens loudness is more reliable than the Logit Lens and than end-of-sentence tokens for finding activations to probe.
\end{itemize}
\section{Related Work}\label{sec:related_work}

\subsection{Probing and its limitations}
We adopt the standard probing setup \citep{belinkov2022probing}, in which probe accuracy is taken as evidence that a property is decodable from a model's representations. Expressive probes may learn the task themselves, which \citet{hewitt2019designing} address by comparing against control tasks, and there is no consensus on probe complexity: some advocate simple probes \citep{alain2016understanding, hewitt2019designing}, others the best-performing one \citep{pimentel2020information}. We therefore report both linear and MLP probes. Finally, high probe accuracy does not entail that the model uses the decoded information \citep{ravichander2021probing, elazar2021amnesic}, although probe-guided interventions can provide causal evidence \citep{giulianelli2018under}. We use probe accuracy only as a measure of decodability, which is what loudness should predict, and make no causal claims.

\subsection{Choosing a layer}
Layers are typically chosen by a heuristic or by an exhaustive sweep. The heuristic view is that middle layers are richest in information, as they sit between layers that process the input and layers that produce the output. Empirically, intermediate layers often yield better representations than final ones \citep{skean2025layer}, and linear probes for spatial and temporal information improve over the early layers before plateauing around the middle of the network \citep{gurnee2024language}. That said, this is only a heuristic. To find definitively which activations work best for a given probing setup, a sweep is often performed. Sweeps train a probe for every layer, and often every position: \citet{lugoloobi2026llms}, for example, cross-validate a linear probe for every layer--position pair to select the layer, position, and regularisation strength for each target. Heuristics are cheap but not signal-specific, while sweeps are signal-specific but costly, especially on long horizon tasks or long reasoning chains.

\subsection{Choosing token positions}
Token-position heuristics fall into three families. The first uses a single fixed position, such as the final input token \citep{marks2024geometry} or a structurally defined position such as the end of the instruction or of the reasoning \citep{arghal2026behavioural}. The second uses positions tied to the signal, such as the last token of an entity name \citep{gurnee2024language} or the exact-answer tokens \citep{orgad2025llms}; relatedly, \citet{tigges2023linear} find sentiment partly summarised at punctuation rather than content tokens. The third aggregates over positions, either with a fixed rule, such as averaging per-token probe scores over the response \citep{goldowsky2025detecting}, or adaptively, such as with attention-based or max-style probes \citep{mckenzie2026detectinghighstakesinteractionsactivation, kramar2026building}. Each heuristic encodes a prior about where a signal is decodable, and all are reliant on the results of narrow-use probes.

\subsection{Label-free readouts of intermediate activations}
A separate line of work reads intermediate activations directly in vocabulary space, without training on labelled data.

\textbf{The Logit Lens} \citep{nostalgebraist2020logitlens} applies the model's unembedding matrix $W_U$ directly to intermediate activations. It is informative in late layers but often uninterpretable in early ones.

\textbf{The J-Lens} \citep{gurnee2026verbalizable} does not unembed an intermediate activation directly. It first estimates how a small change to the activation at layer $l$ changes the model's final layer, using the Jacobian between the two layers averaged over many positions and inputs, $\bar J_l$. Multiplying an activation $a_{l,p}$ by $\bar J_l$ translates it into the final layer's space, where the unembedding applies:
\begin{equation}
  \mathrm{JLens}_l(a_{l,p}) = \mathrm{softmax}\big(W_U \bar J_l\, a_{l,p}\big).
\end{equation}
It recovers the tokens a model is poised to verbalize at earlier depths than the Logit Lens. We fit our own J-Lens for each model (Appendix~\ref{appx:details}).

These lenses have so far been used to inspect what an activation encodes; we use them to predict where a probe can decode a signal.

\section{Methods}\label{sec:methods}

\begin{wrapfigure}{r}{0.45\linewidth}
  \vspace{-16pt}
  \centering
  \includegraphics[width=\linewidth]{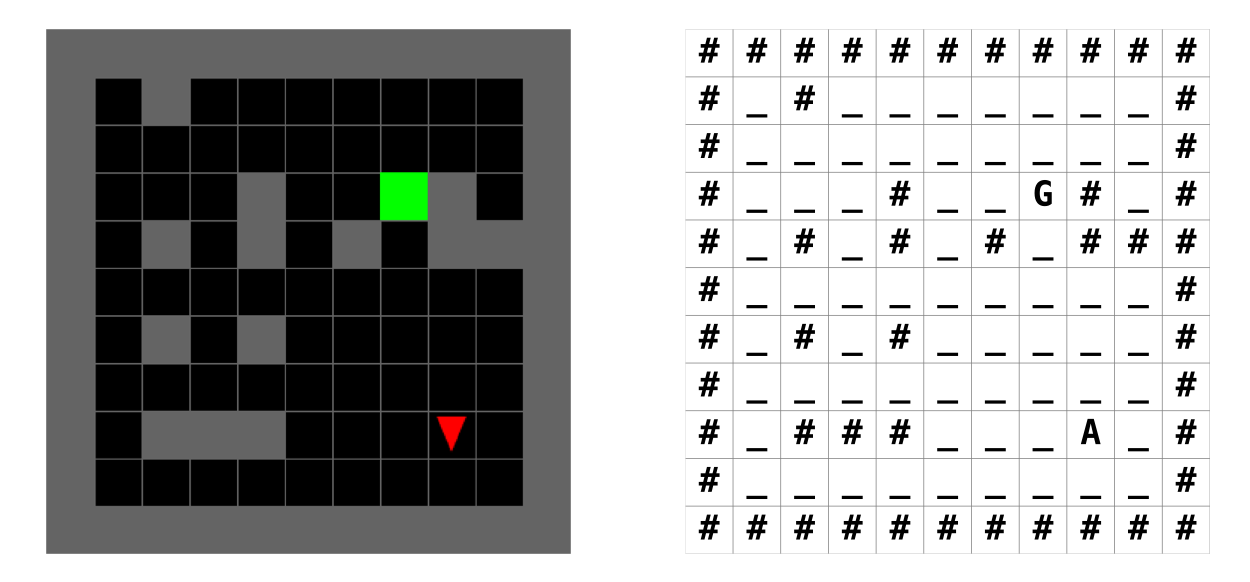}
  \caption{An example grid (left) and its text representation in the prompt (right). Reproduced from \citet[Figure~3]{arghal2026behavioural}.}
  \label{fig:grid_env}
  \vspace{-4pt}
\end{wrapfigure}
\paragraph{Environment and data.}
We study GPT-OSS-20B and Qwen3.6-35B-A3B in the grid-world environment studied by \citet{arghal2026behavioural}, without doors or keys. At each step, the model controls an agent \texttt{A} on an $n\times n$ grid containing walls \texttt{\#}, empty cells \texttt{\_} and a goal \texttt{G} (Figure~\ref{fig:grid_env}); it produces a reasoning chain before producing a final answer indicating the action the agent will take (\texttt{UP}, \texttt{DOWN}, \texttt{LEFT} or \texttt{RIGHT}) in a separate output channel. Like \citet{arghal2026behavioural}, we use grid sizes $n \in \{5,7,\dots,15\}$, complexity levels $c \in \{0.0,0.2, \dots, 1.0\}$, and treat the reasoning chain of each step as an independent example and split training and evaluation data by chain. Because Qwen3.6-35B-A3B produces much longer chains, we uniformly sample $20\%$ of its reasoning-token positions before any other selection or analysis takes place.

\paragraph{Signals and vocabularies.}
An activation encodes information about many aspects of the model's input and computation at once. We call the information about one such aspect a \emph{signal}. Equivalently, a signal can be defined as a set of related features; for direction, these include \emph{left}, \emph{right}, \emph{up}, \emph{down}, \emph{horizontal}, and \emph{vertical} to name a few examples. A signal is not tied to a particular set of labels: many probing tasks can target the same signal, each reading out different features with its own classes. For instance, one task may predict which of the four directions the agent moves next, and another only whether that move is horizontal or vertical. Yet another probe may be trained to predict the general direction the goal state may be in, the list of examples goes on.  We say that a signal is decodable for a task at $(l,p)$ if a probe trained on $a_{l,p}$ predicts the task's labels above chance. Loudness, in contrast, depends only on the signal, the layer and the position.
In this study, we focus on the movement \emph{direction} signal $d$. We probe it with one task, the model's \emph{local belief} at token position $p$: we truncate the reasoning chain at $p$, close the reasoning channel and open the final-answer channel. This forces the model to output an action, which we subsequently take as our local belief label. Because this label is derived from the model's own output, we add a second task for the same signal whose label comes from the environment: the \emph{goal side}, i.e.\ whether the goal lies to the left or to the right of the agent, read from the grid coordinates (steps where both share a column are dropped; chance is $0.5$). The \emph{grid} signal $g$, corresponding to the layout of the grid, serves as a control. Probes associated with this signal are trained to predict the content of a queried cell. For each model and signal, we construct the vocabulary $V_s$ from words that express the signal rather than any task's classes: a small set of anchor words which we expand in two stages: a lexical matching stage, and a semantic matching stage. This method is explained in depth in Appendix~\ref{appx: capturing_signals}, however, we do not intend our vocabulary building method to be presented as definitive. Indeed, it results in many false positives and we expect future work to surface methods much better at capturing a relevant signal of interest. 

\paragraph{J-Lens loudness.}
We compute loudness with the J-Lens $\mathrm{JLens}_l$ that we fit for each model (Section~\ref{sec:related_work}; Appendix~\ref{appx:details}). For a signal $s$ with vocabulary $V_s$, J-Lens loudness is the log-probability mass assigned to $V_s$:
\begin{equation}
  J_s(a_{l,p}) = \log \sum_{v \in V_s} \mathrm{JLens}_l(a_{l,p})_v .
\end{equation}
Logit-Lens loudness is defined identically with $\bar J_l$ replaced by the identity, i.e.\ with the Logit Lens in place of the J-Lens. We select the loudest layer for each signal and, within each reasoning chain $x$, the $N$ loudest positions at that layer:
\begin{equation}
  l^*_s = \operatorname*{arg\,max}_{l} \; \mathbb{E}_{x \sim \mathcal{D}} \, \mathbb{E}_{p} \big[ J_s(a_{l,p}) \big],
  \qquad
  P^*_{s,N}(x) = \operatorname*{top\text{-}N}_{p} \; J_s(a_{l^*_s,p}).
\end{equation}
Neither step uses probe labels or probe training.

\paragraph{Position selection.}
In addition to J-Lens loudness, we also compute Logit-Lens loudness at $l^*_s$ with the same $V_s$, so the two loudness metrics differ only in the mapping to vocabulary space. We compare two selection schemes. \emph{Top-$N$} keeps the $N$ positions per chain with the highest J-Lens loudness, the highest Logit-Lens loudness, or chosen uniformly at random ($N{=}20$ for GPT-OSS-20B and $N{=}60$ for Qwen3.6-35B-A3B, reflecting chain length). \emph{Sentence-level} selection keeps one position per sentence: the loudest under each lens, a random one, or the sentence-final punctuation token (EOS), a structural heuristic for where information is summarised \citep{tigges2023linear}. We apply sentence-level selection to GPT-OSS-20B only, as the reasoning chains of Qwen3.6-35B-A3B are too long to process every sentence under our computational constraints.

\paragraph{Probes.}
We train linear and MLP probes on activations at $l^*_s$, one per model, signal, training criterion and architecture. The direction probe predicts one of the four actions. The grid probe receives the activation concatenated with the coordinates of the queried cell and predicts \texttt{wall}, \texttt{empty}, \texttt{goal} or \texttt{agent} (Appendix~\ref{appx:grid_results}). To validate the layer selection, we additionally train direction and goal-side probes at every layer on the same top-$N$ J-Lens and random positions, which are ranked once at $l^*_d$ (5 seeds for GPT-OSS-20B, 3 for Qwen3.6-35B-A3B). As a label-free baseline, we also use the J-Lens directly as a classifier, predicting the action whose token it rates highest. Training details are in Appendix~\ref{appx:details}.

\paragraph{Evaluation.}
On held-out chains, we construct every selection subset independently and evaluate each probe on all of them. This cross-selection evaluation separates whether a criterion provides good training data from whether it finds activations where the signal is decodable. We report balanced accuracy throughout. To test whether decodability grows with loudness, we pool all held-out tokens, bin them into loudness deciles and compute accuracy per decile, with $95\%$ bootstrap confidence intervals over chains ($300$ resamples).

It is possible that the loudest tokens are verbalized tokens of $V_s$, that is, tokens the model actually writes in its reasoning chain. For example, the loudest tokens for direction might be \texttt{.left}, \texttt{derecha} or \texttt{$\rightarrow$}, all of which are in $V_d$. We would like loudness to find more than these trivial positions, so in an additional analysis (Appendix Figure~\ref{fig:appx_decile_nowords}) we remove every verbalized token of $V_s$, as well as the tokens within $\pm2$ positions of it, and recompute the deciles.

To test selectivity, we also test how the same direction probes perform along the grid loudness axis.
\section{Results}
\label{sec:results}

We first examine where loudness peaks across layers (Section~\ref{sec:results_layers}). At the selected layer, we then test whether loud positions are more decodable than Logit-Lens-selected and random ones (Section~\ref{sec:results_topn}), whether decodability increases with loudness (Section~\ref{sec:results_graded}), and whether this relationship is specific to the signal (Section~\ref{sec:results_selectivity}). We then repeat the analysis with labels taken from the environment (Section~\ref{sec:results_goal_side}). Direction is our main signal; the grid signal serves as a control (Section~\ref{sec:results_grid}). All scores are balanced accuracy; chance is $0.25$ unless stated otherwise.

\subsection{Loudness peaks in an intermediate layer band}
\label{sec:results_layers}
\begin{figure}[h]
  \centering
  \begin{tabular}{@{}c@{\hspace{2pt}}cc@{}}
    & \small Direction & \small Grid \\
    \rotatebox{90}{\small\hspace{1.2em}GPT-OSS-20B} &
    \includegraphics[width=0.47\linewidth]{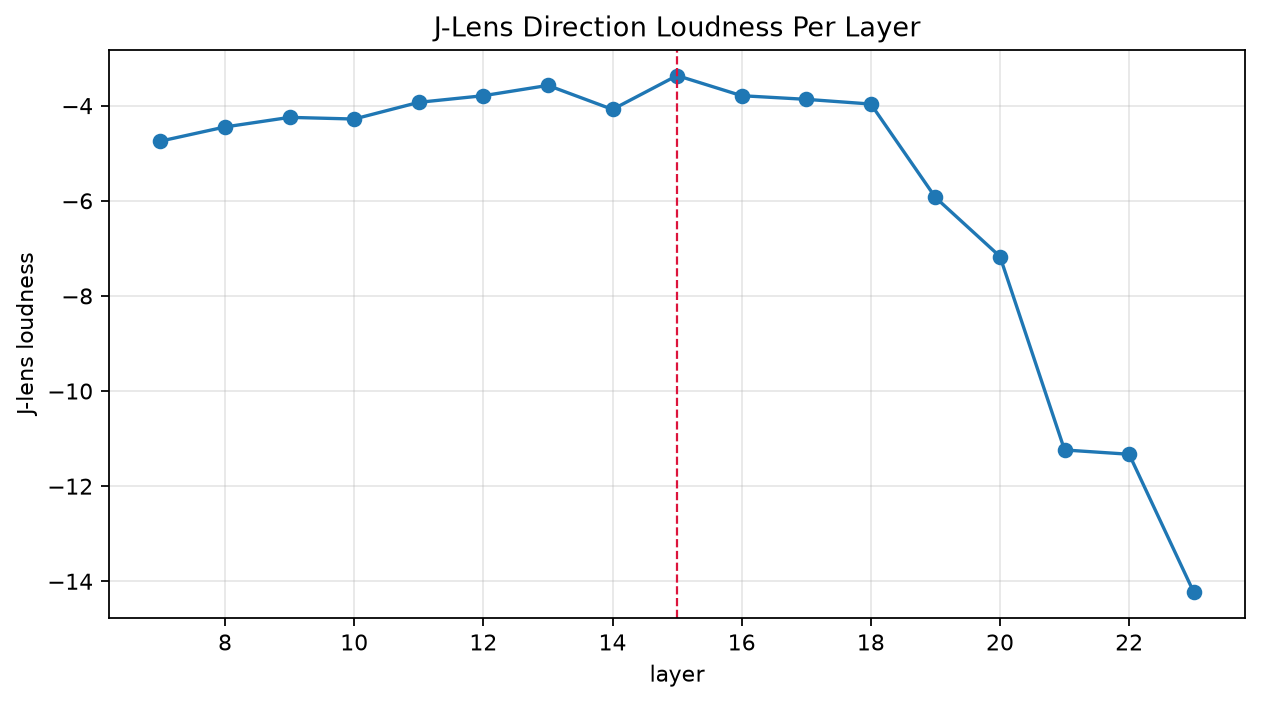} &
    \includegraphics[width=0.47\linewidth]{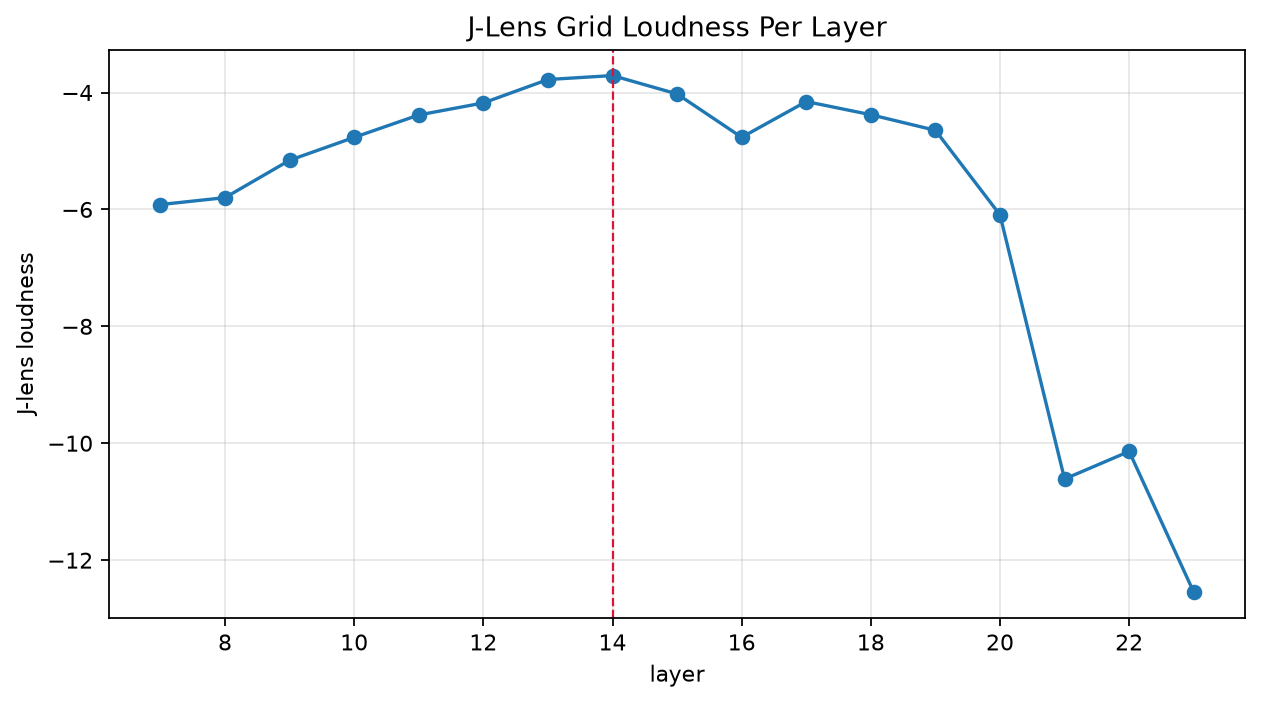} \\
    \rotatebox{90}{\small\hspace{0.8em}Qwen3.6-35B-A3B} &
    \includegraphics[width=0.47\linewidth]{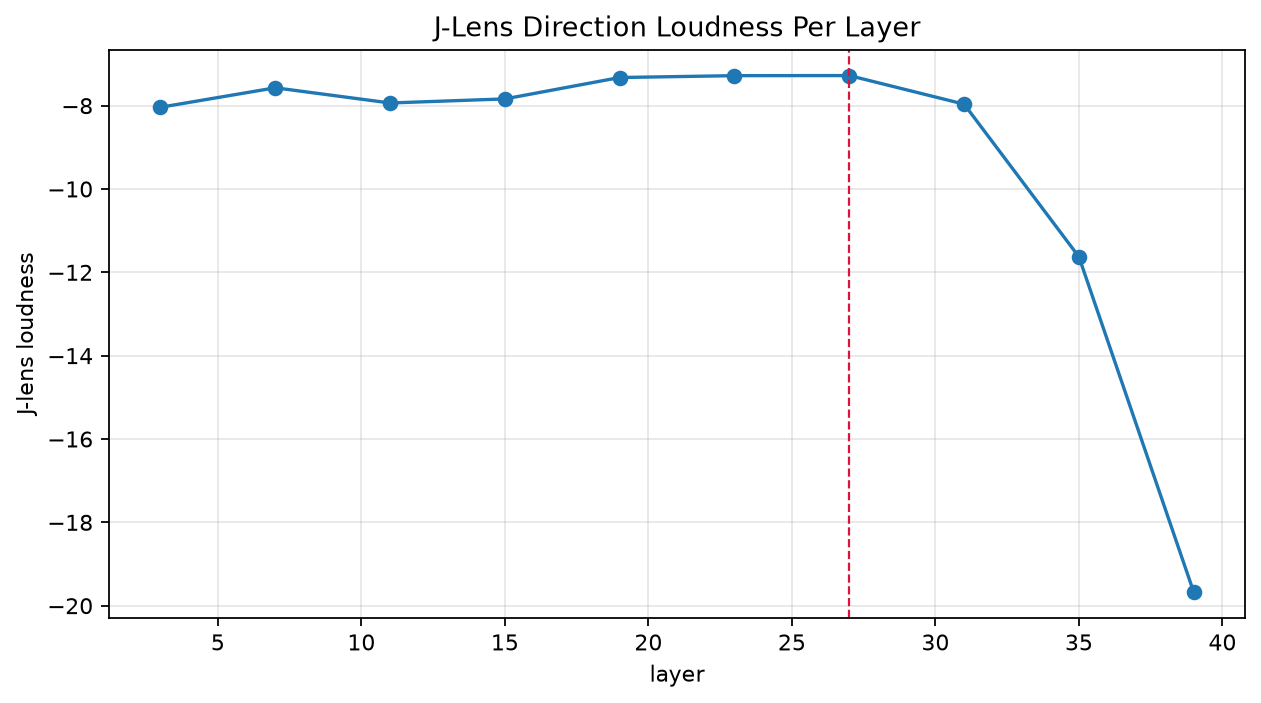} &
    \includegraphics[width=0.47\linewidth]{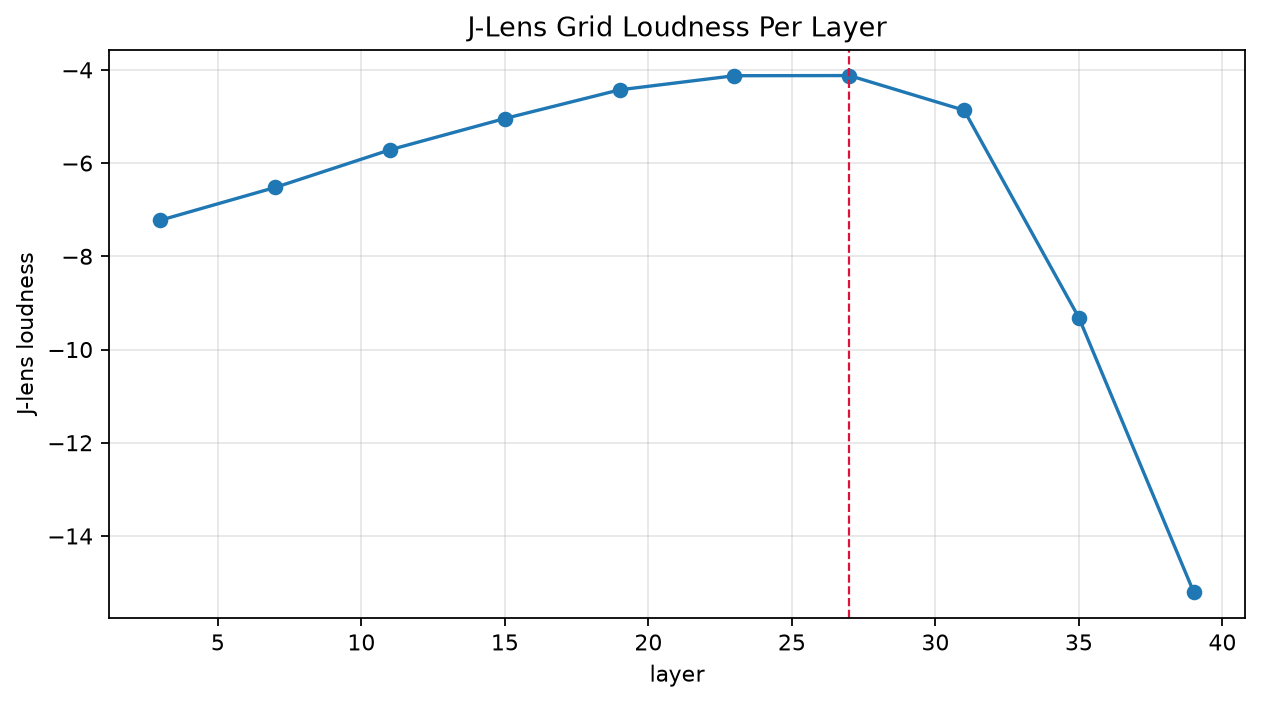} \\
  \end{tabular}
  \caption{Mean J-Lens loudness per layer. Dashed line: selected layer $l^*_s$.}
  \label{fig:layer_profiles}
\end{figure}

Averaged over tokens and reasoning chains, J-Lens loudness was high across a broad intermediate band and fell sharply in the final layers, for both models and both signals (Figure~\ref{fig:layer_profiles}). It peaked at layer $15$ (direction) and $14$ (grid) for GPT-OSS-20B, and at layer $27$ for both signals in Qwen3.6-35B-A3B. This band matches where probe accuracy typically peaks in prior work \citep{skean2025layer, gurnee2024language, lugoloobi2026llms}. Logit-Lens profiles were less informative and, for Qwen3.6-35B-A3B, flat across nearly all layers (Appendix Figure~\ref{fig:appx_logitlens_layers}). We therefore rank positions at the J-Lens-selected layer for every criterion. The selected layer does not depend on the exact word list: removing a random $20\%$ of $V_d$ (five draws, GPT-OSS-20B) never moved the argmax from layer $15$, and every perturbed profile correlated with the full one at $r \geq 0.9999$ (Appendix Figure~\ref{fig:appx_vocab_ablation}).

\begin{figure}[t]
  \centering
  \includegraphics[width=\linewidth]{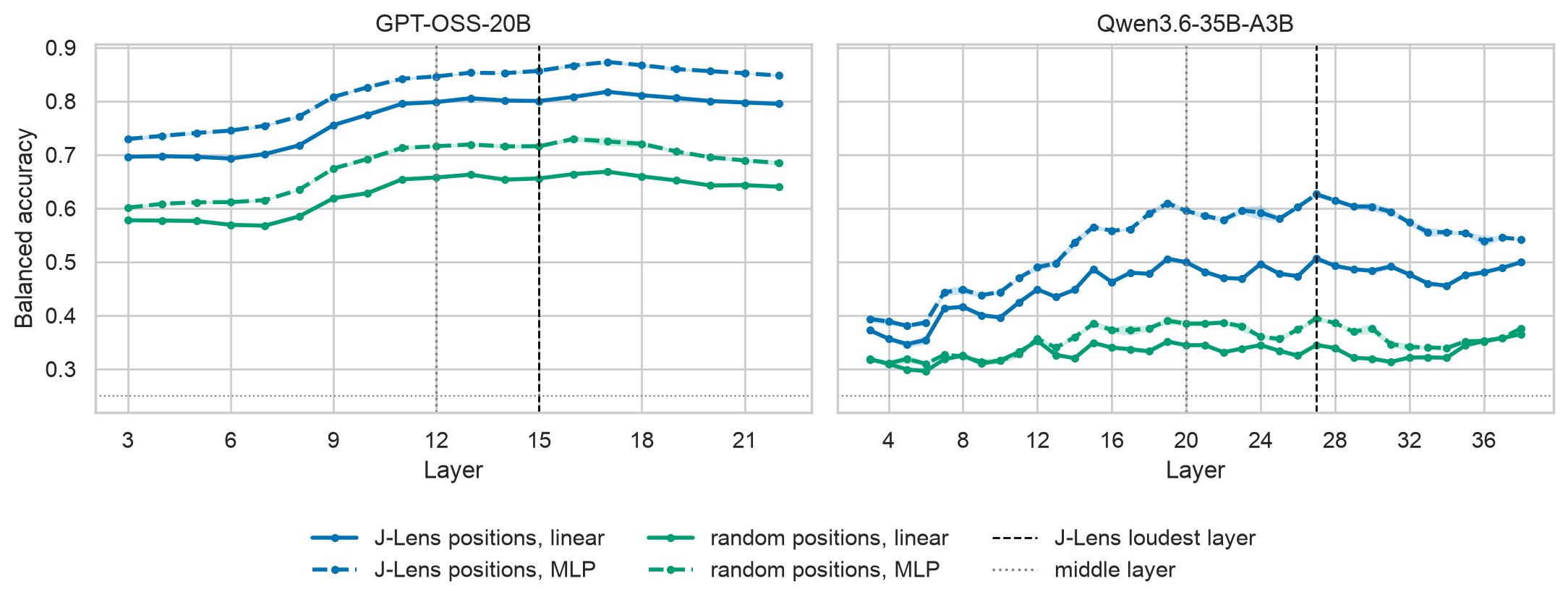}
  \caption{Direction probes trained and evaluated at every layer, on the top-$N$ J-Lens positions (ranked once at $l^*_d$) and on random positions. Bands: $\pm1$ s.d.\ over seeds. Dashed: J-Lens loudest layer; dotted: middle layer; horizontal: chance.}
  \label{fig:layer_sweep}
\end{figure}

\paragraph{Validation against a layer sweep.}
A probe sweep over all layers (Figure~\ref{fig:layer_sweep}; Appendix Tables~\ref{tab:appx_sweep_gptoss} and~\ref{tab:appx_sweep_qwen}) shows that the loudest layer lies on the plateau of highest decodability, but is not always its maximum. For Qwen3.6-35B-A3B, layer $27$ was the most decodable layer for both probes ($.507$ linear, $.626$ MLP; layer $19$ is level for linear probes). For GPT-OSS-20B, accuracy barely changed between layers $11$ and $20$; the best layer was $17$ ($.818$ / $.873$), $0.016$--$0.017$ above the loudest layer $15$ ($.800$ / $.857$), a gap larger than the seed s.d.\ ($\le 0.005$). For this signal, loudness therefore does not clearly improve on the middle-layer heuristic. Positions matter more than layers: loud positions were more decodable than random ones at \emph{every} layer, by $0.12$--$0.16$ for GPT-OSS-20B and up to $0.25$ for Qwen3.6-35B-A3B, even though they were ranked at a single layer. Finally, the J-Lens alone does not decode the direction: used as a classifier, it is close to chance at the loudest layers ($.266$ for GPT-OSS-20B at layer $15$, $.332$ for Qwen3.6-35B-A3B at layer $27$; Appendix Figure~\ref{fig:appx_lens_classifier}). The lens indicates \emph{where} the information is present; a probe is needed to extract it.
\subsection{Loud positions are more decodable}
\label{sec:results_topn}

\begin{table}[h]
  \centering
  \caption{Direction, top-$N$ selection. Rows: training criterion; columns: evaluation subset. \textbf{Bold}: best subset per probe; \underline{underlined}: in-distribution.}
  \label{tab:topn_dir}
  \small
  \setlength{\tabcolsep}{4.5pt}
  \begin{tabular}{llcccccc}
    \toprule
    & & \multicolumn{3}{c}{GPT-OSS-20B} & \multicolumn{3}{c}{Qwen3.6-35B-A3B} \\
    \cmidrule(lr){3-5}\cmidrule(lr){6-8}
    Trained on & Probe & J-Lens & Logit Lens & Random & J-Lens & Logit Lens & Random \\
    \midrule
    J-Lens     & Linear & \underline{\textbf{.800}} & .689 & .609 & \underline{\textbf{.503}} & .430 & .320 \\
               & MLP    & \underline{\textbf{.860}} & .745 & .655 & \underline{\textbf{.621}} & .500 & .346 \\
    Logit Lens & Linear & \textbf{.759} & \underline{.726} & .620 & \textbf{.476} & \underline{.450} & .342 \\
               & MLP    & \textbf{.818} & \underline{.790} & .676 & \textbf{.598} & \underline{.547} & .372 \\
    Random     & Linear & \textbf{.751} & .698 & \underline{.655} & \textbf{.432} & .421 & \underline{.347} \\
               & MLP    & \textbf{.818} & .765 & \underline{.721} & \textbf{.528} & .469 & \underline{.396} \\
    \bottomrule
  \end{tabular}
\end{table}

Table~\ref{tab:topn_dir} evaluates every probe on held-out positions selected by each criterion, with $14{,}391$ (GPT-OSS-20B) and $3{,}120$ (Qwen3.6-35B-A3B) positions per subset. In both models, every probe was most accurate on J-Lens-selected positions, followed by Logit-Lens-selected and then random positions. This ordering also holds for probes trained on random positions, whose training data is independent of loudness, so it reflects a property of the selected activations rather than of the probe. Training on loud positions gave the highest accuracy on loud positions but generalised less well to random ones than training on random positions. Loudness is therefore most useful for deciding \emph{where to apply} a probe.

\begin{wraptable}{R}{0.5\linewidth}
  \vspace{-25pt}
  \centering
  \caption{GPT-OSS-20B, direction, sentence-level selection: J-Lens/Logit Lens loudest, random uniform, or punctuation token (EOS). \textbf{Bold}: best subset per probe; \underline{underlined}: in-distribution.}
  \label{tab:p1}
  \footnotesize
  \setlength{\tabcolsep}{3pt}
  \resizebox{\linewidth}{!}{%
  \begin{tabular}{llcccc}
    \toprule
    Trained on & Probe & J-Lens & Logit Lens & Random & EOS \\
    \midrule
    J-Lens     & Linear & \underline{\textbf{.606}} & .531 & .460 & .460 \\
               & MLP    & \underline{\textbf{.677}} & .605 & .513 & .542 \\
    Logit Lens & Linear & .565 & \underline{\textbf{.571}} & .474 & .463 \\
               & MLP    & \textbf{.642} & \underline{.639} & .535 & .542 \\
    Random     & Linear & \textbf{.570} & .551 & \underline{.509} & .518 \\
               & MLP    & \textbf{.648} & .618 & \underline{.584} & .606 \\
    EOS        & Linear & .489 & .446 & .413 & \underline{\textbf{.580}} \\
               & MLP    & .530 & .481 & .445 & \underline{\textbf{.657}} \\
    \bottomrule
  \end{tabular}}
  \vspace{-1pt}
\end{wraptable}
\paragraph{Comparison with a fixed-position heuristic.}
Selecting one position per sentence allows a comparison with end-of-sentence punctuation (EOS), a position at which information is thought to be summarised \citep{tigges2023linear} (Table~\ref{tab:p1}; GPT-OSS-20B). The top-$N$ pattern holds at this granularity: every probe not trained on EOS positions was more accurate on J-Lens-selected than on EOS positions, and J-Lens probes evaluated in-distribution outperformed EOS probes evaluated in-distribution. EOS probes were accurate only on EOS positions and transferred poorly elsewhere.

\subsection{Decodability increases with loudness}
\label{sec:results_graded}

\begin{figure}[t]
  \centering
  \begin{tabular}{@{}c@{\hspace{2pt}}cc@{}}
    & \small Ranked by J-Lens loudness & \small Ranked by Logit-Lens loudness \\
    \rotatebox{90}{\small\hspace{1.6em}GPT-OSS-20B} &
    \includegraphics[width=0.47\linewidth]{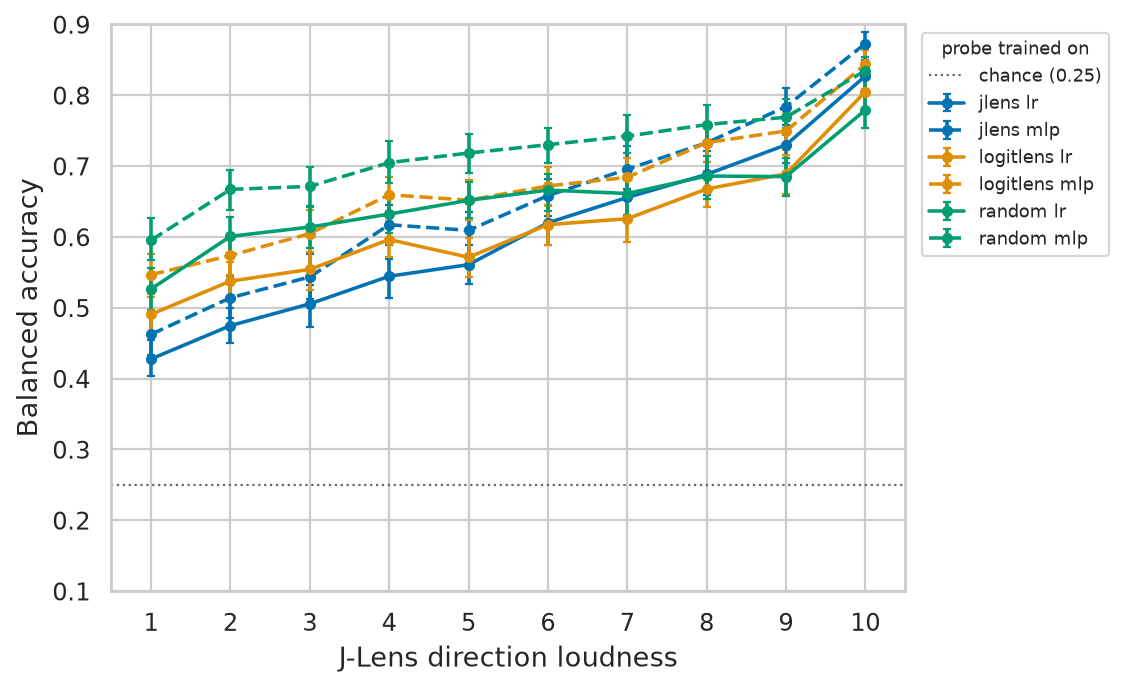} &
    \includegraphics[width=0.47\linewidth]{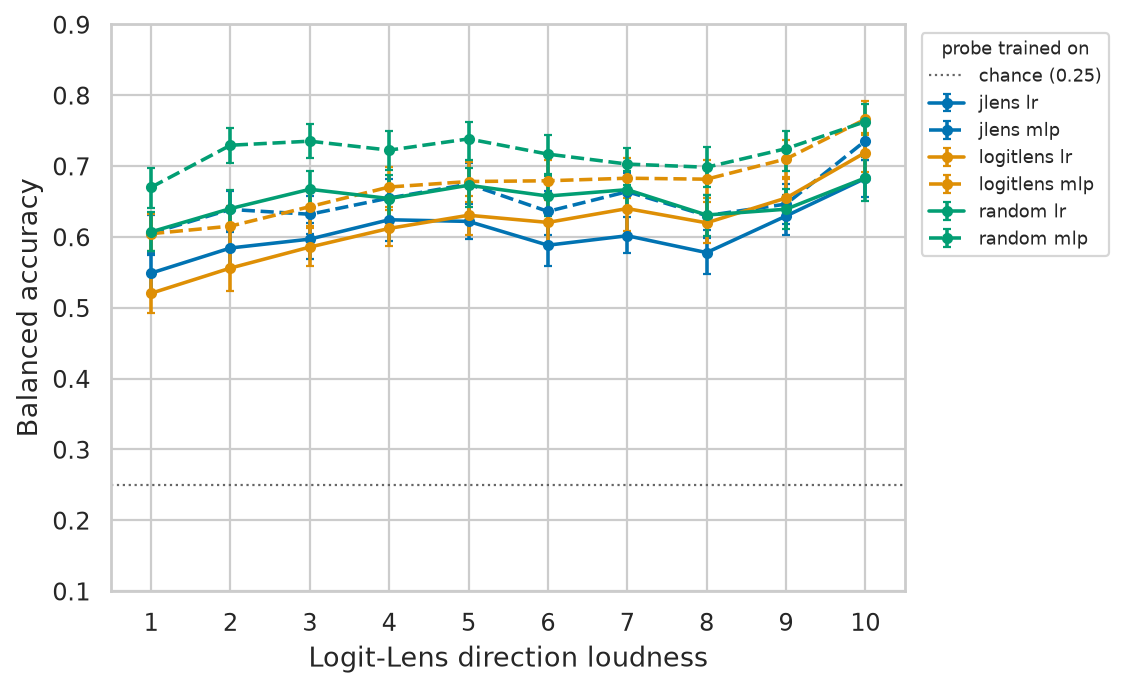} \\
    \rotatebox{90}{\small\hspace{1.2em}Qwen3.6-35B-A3B} &
    \includegraphics[width=0.47\linewidth]{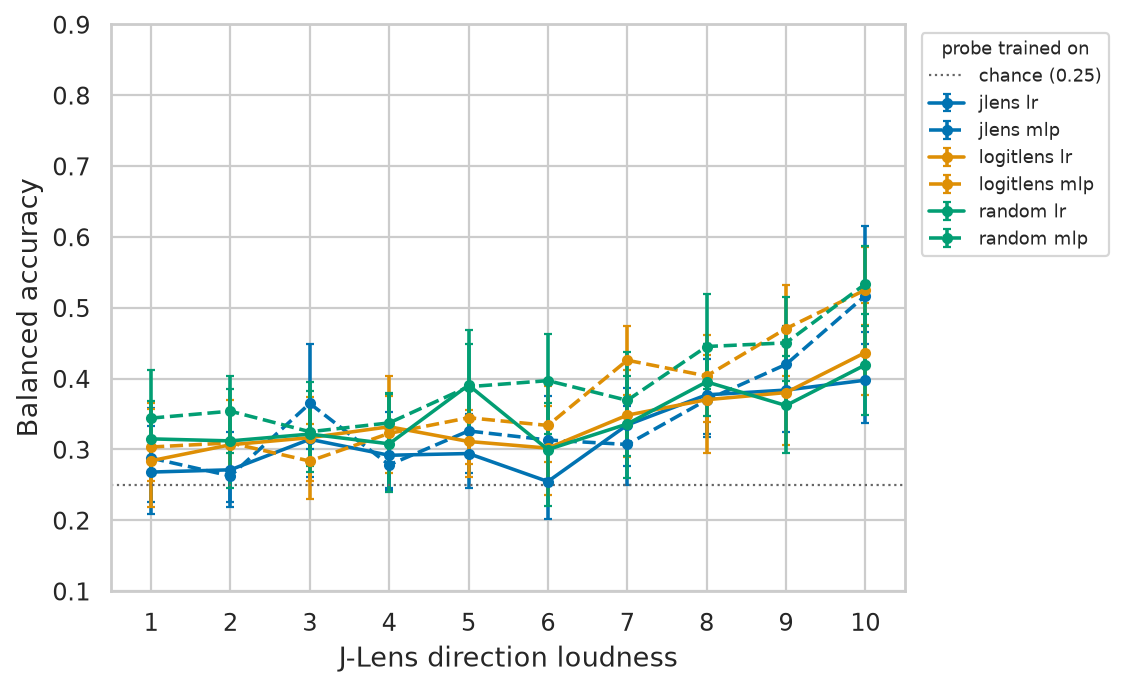} &
    \includegraphics[width=0.47\linewidth]{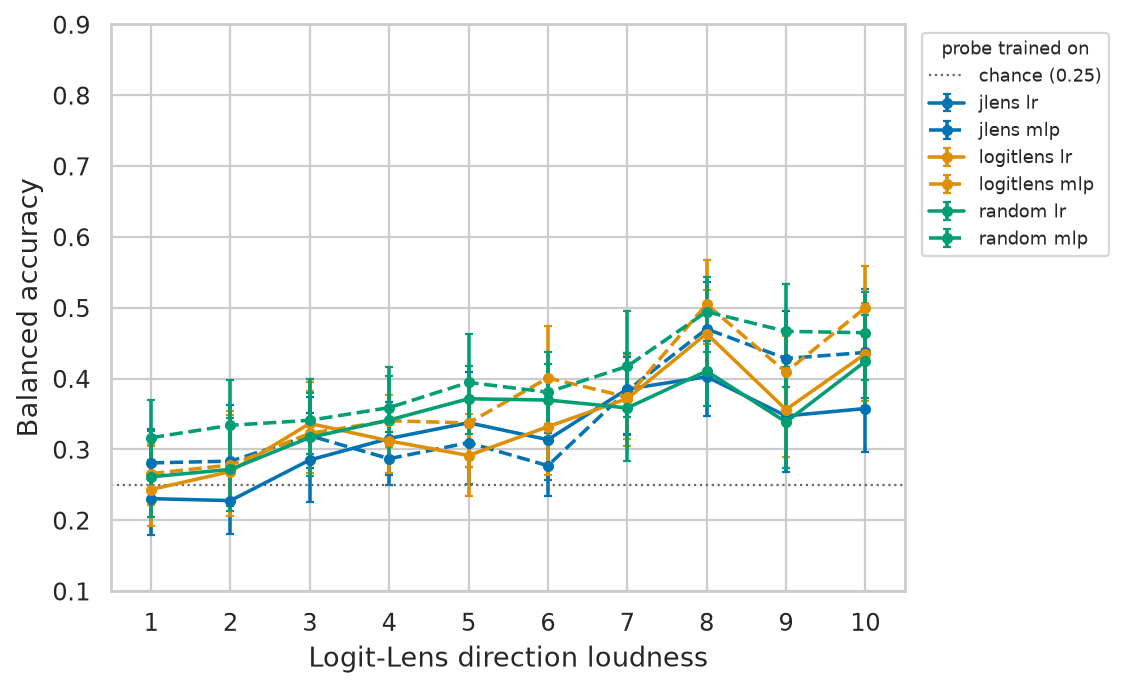} \\
  \end{tabular}
  \caption{Direction probes by direction-loudness decile at $l^*_d$ (1: quietest, 10: loudest). Line label: training criterion; solid: linear, dashed: MLP. Error bars: 95\% bootstrap CIs.}
  \label{fig:deciles_direction}
\end{figure}

Binning all held-out tokens by loudness, with each line a probe labelled by its training criterion, shows that the top-$N$ result is part of a graded relationship (Figure~\ref{fig:deciles_direction}). For GPT-OSS-20B, accuracy increased almost monotonically with J-Lens loudness for every probe, whereas Logit-Lens loudness ordered tokens poorly outside the loudest decile. For Qwen3.6-35B-A3B, which has fewer held-out chains and wider intervals, accuracy rose over the loudest deciles under both lenses.

A possible confound is that loud tokens are simply direction words (e.g.\ \texttt{left}), which a probe could decode trivially. After removing every token of $V_d$ together with the two tokens on either side from the list of selected tokens, the J-Lens trend persisted in both models, whereas the Logit-Lens trend for GPT-OSS-20B disappeared (Appendix Figure~\ref{fig:appx_decile_nowords}). J-Lens loudness therefore tracks decodability away from the trivial positions where direction words are written, and can identify positions where the signal is decodable beyond its surface form.

An additional look into which tokens are selected can be found in Appendix Section \ref{appx:selected_tokens}, where we find the models' direction-related computations seem to also take place during the planning of future grid states, or when recalling the agent's or goal's grid locations. 
\subsection{Loudness is signal-specific}
\label{sec:results_selectivity}

\begin{figure}[t]
  \centering
  \includegraphics[width=0.47\linewidth]{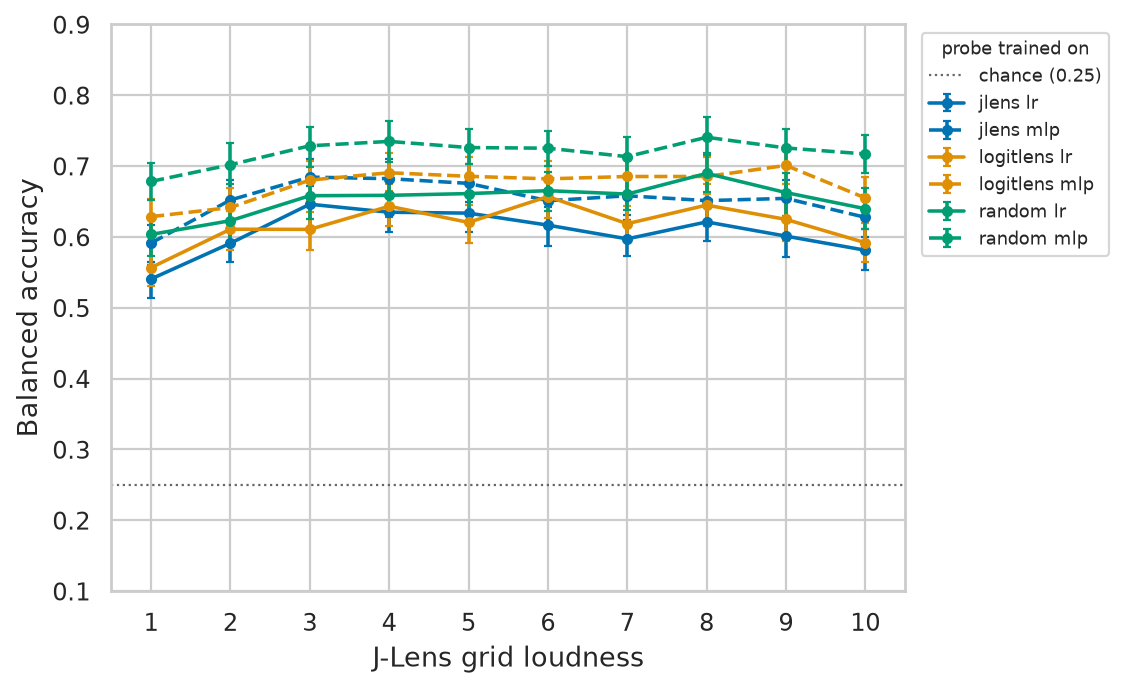}
  \includegraphics[width=0.47\linewidth]{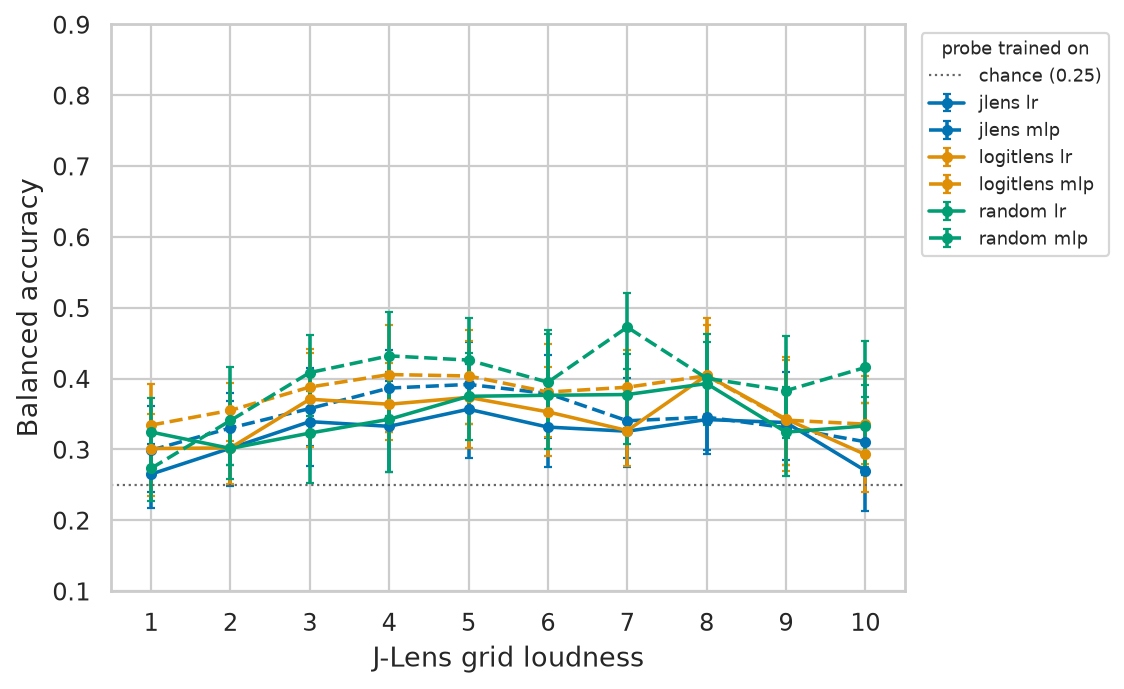}
  \caption{Direction probes by J-Lens \emph{grid}-loudness decile. Left: GPT-OSS-20B; right: Qwen3.6-35B-A3B.}
  \label{fig:selectivity}
\end{figure}

If loudness merely marked generically informative positions, the loudness of an unrelated signal would predict direction decodability equally well. It does not. Direction and grid J-Lens loudness were nearly uncorrelated across tokens (Spearman $\rho \approx 0.04$ in both models; Appendix Figure~\ref{fig:appx_signal_vs_signal}), and binning the direction probes of Figure~\ref{fig:deciles_direction} by grid loudness instead produced no consistent trend (Figure~\ref{fig:selectivity}).

\subsection{Loudness predicts decodability of environment labels}
\label{sec:results_goal_side}

\begin{figure}[t]
  \centering
  \includegraphics[width=0.47\linewidth]{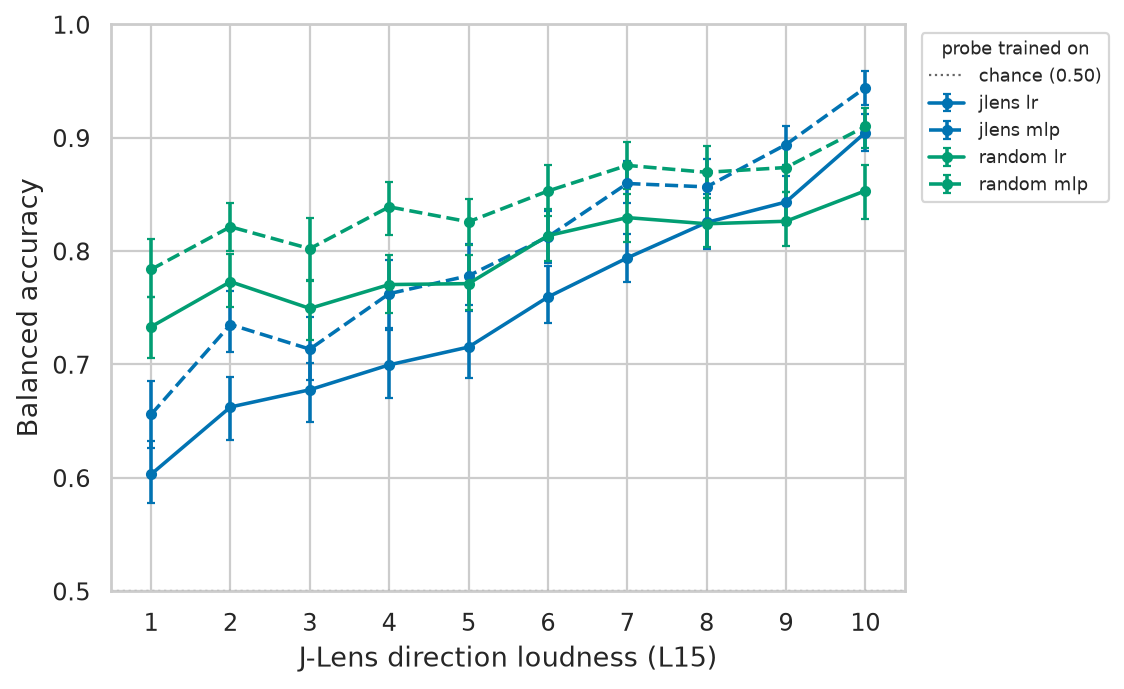}
  \includegraphics[width=0.47\linewidth]{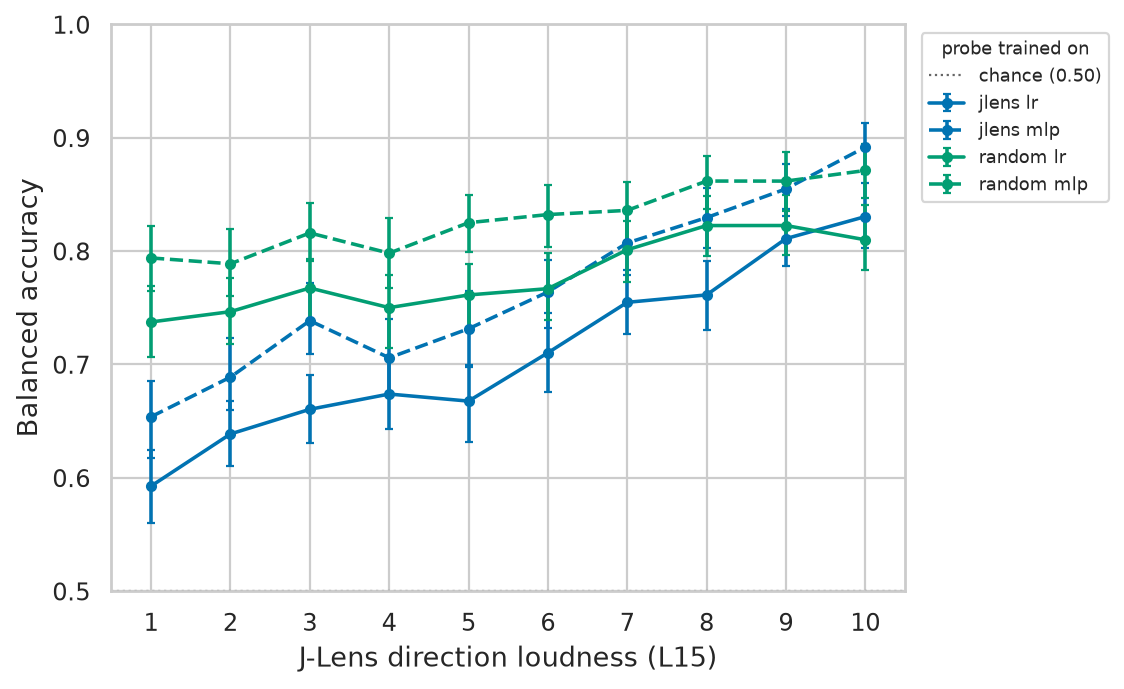}
  \caption{GPT-OSS-20B goal-side probes by J-Lens direction-loudness decile at $l^*_d$, on held-out random positions. Left: all tokens; right: direction words and their $\pm2$ neighbours removed. Chance is $0.5$. Conventions as in Figure~\ref{fig:deciles_direction}.}
  \label{fig:goal_side_deciles}
\end{figure}

The local-belief label is the model's own forced answer, so high accuracy at loud positions could partly reflect what the model is about to say. The goal-side label (Section~\ref{sec:methods}) is instead read from the grid. It behaved like the direction label. In the layer sweep (Appendix Figure~\ref{fig:appx_goal_sweep}), probes on loud positions beat probes on random positions at every layer, by $0.06$--$0.10$ for GPT-OSS-20B and up to $0.19$ for Qwen3.6-35B-A3B. The best layers were again on the plateau around the loudest layer (GPT-OSS-20B: layer $17$, $.907$ linear / $.937$ MLP, against $.884$ / $.927$ at layer $15$; Qwen3.6-35B-A3B: layer $27$ for MLP probes, $.804$). Binned by J-Lens direction loudness, the accuracy of GPT-OSS-20B goal-side probes rose from $.60$ to $.91$ (linear) and $.66$ to $.94$ (MLP) for probes trained on loud positions, and the trend persisted after removing direction words (Figure~\ref{fig:goal_side_deciles}). Binned by grid loudness, it was flat ($|\Delta| \le 0.02$ between the extreme deciles; Appendix Figure~\ref{fig:appx_goal_deciles}). Because the goal side correlates with \texttt{LEFT}/\texttt{RIGHT} actions, this task does not fully separate the goal from the plan, but it removes the dependence of the label on the model's output.

\subsection{The grid signal as a control}
\label{sec:results_grid}

We also trained probes to decode the grid state directly. These probes performed poorly on every position subset: linear probes stayed near chance and MLP probes only modestly above it. With so little decodable signal, differences between selection criteria were small and inconsistent across models and lenses (Appendix~\ref{appx:grid_results}). We therefore treat the grid experiment as inconclusive about grid decodability and use grid loudness only as a control signal. We discuss the design choices behind this outcome in Section~\ref{sec:discussion}.

\section{Discussion}
\label{sec:discussion}

Our results show that J-Lens loudness, computed without labels or probe training, predicts where the model's current movement decision is decodable in the reasoning chains of two models. Loud positions are more decodable than randomly selected ones for every direction probe we trained, decodability increases steadily with loudness, and the relationship survives removing the positions where direction words are written. Loudness for an unrelated signal carries no such information, indicating that the score is specific to the signal whose vocabulary defines it.

\paragraph{Why the J-Lens outperforms the Logit Lens.}
Both lenses read an intermediate activation in vocabulary space, but the Logit Lens applies the final unembedding directly to a representation that is not yet in the final layer's space. In intermediate layers it is therefore poorly calibrated: for Qwen3.6-35B-A3B, its loudness is flat across nearly all layers and varies within a narrow band across tokens (Appendix Figures~\ref{fig:appx_logitlens_layers} and~\ref{fig:appx_lens_vs_lens}). The J-Lens first maps the activation to the final layer through the model's average Jacobian and so reflects what the model is poised to verbalize \citep{gurnee2026verbalizable}. This distinction matters most away from the surface text. When direction words and their neighbours are removed, Logit-Lens loudness no longer tracks decodability in GPT-OSS-20B, whereas J-Lens loudness still does. This suggests that J-Lens loudness captures latent content about the decision rather than the identity of nearby tokens, which builds onto the interpretation of the J-Space as a global workspace \citep{gurnee2026verbalizable}.

\paragraph{Practical use and cost.}
Loudness requires a signal vocabulary and a J-Lens for the layers of interest. Fitting the J-Lens is the most expensive step of the method, but it is done once per model, needs no task labels, is reused across signals, and is not needed at all when a fitted lens is already available. Given a lens, ranking positions costs one forward pass over the chains and one lens readout per token, comparable to applying a probe at every token, with no labels and no probe training. A probe sweep instead trains a probe for every candidate layer (and, for positions, every candidate position), and needs a label at every candidate; for the local-belief label, each label is a separate forced generation. When the layer is fixed by the middle-layer heuristic, both approaches train a single probe, and the difference lies at inference: loudness is computed in the same forward pass as the probe and, since accuracy rises with loudness (Figure~\ref{fig:deciles_direction}), it indicates how far each per-token probe output can be trusted. Our cross-selection results also suggest a division of labour: probes trained on broadly sampled positions generalise better, while loud positions are where any probe is most accurate. Loudness is therefore best used to decide where to apply a probe, and where to look for a signal in the first place.

\paragraph{Why the grid signal was inconclusive.}
The grid signal was intended as a second test of the method, but our design choices prevented it from serving that role. First, the probing task may be too broad in scope. The grid probe asks for the content of any queried cell, given the cell's coordinates concatenated to the activation, so a linear probe cannot condition its reading on the queried cell, and both linear and MLP probes decoded little of the grid state. Moreover, the model has little reason to hold the entire layout at every reasoning token, since it can attend back to the grid in the prompt. A narrower task, such as predicting whether each cell adjacent to the agent is a wall, would target information the model plausibly computes and carries forward for its next move, and would use coordinates relative to the agent, which mean the same thing across grid sizes. Second, the signal itself may be too broad. The grid state comprises the content of every cell, and grid loudness is a single score over a pooled vocabulary of cell-content words. Such a score measures whether the model is processing the grid in general rather than whether a particular cell is represented, and expanding generic anchors such as \texttt{wall} or \texttt{goal} likely admitted tokens used outside descriptions of the current grid. Loudness, as it is currently designed, seems to therefore be more likely to accurately predict decodability for signals smaller in scope. We treat the grid results as inconclusive, but believe decodability prediction is an underexplored problem which J-Lens loudness pushes forward.

\paragraph{Limitations.}
Beyond the grid signal, we study one environment and one decodable signal in two models, probed with two tasks. The goal-side task removes the dependence of the label on the model's output, but it correlates with the next action, so it does not isolate the goal from the plan. The layer sweep shows that the loudest layer lies on the high-decodability plateau, but not that it is the best layer, and for Qwen3.6-35B-A3B loudness was profiled only at the nine layers with a fitted J-Lens. We aim to improve the paper in the next iteration by extending our analysis to more signals, more models, and more environments.

\section{Conclusion}
\label{sec:conclusion}

We introduced J-Lens loudness, the log-probability mass that the J-Lens assigns to a signal's vocabulary, as a label-free criterion for choosing where to apply a probe. In the reasoning chains of GPT-OSS-20B and Qwen3.6-35B-A3B, positions that were loud for the agent's movement direction were more decodable than positions selected by Logit-Lens loudness, at random, or at sentence-final punctuation. Probe performance seems to increase steadily with loudness, which indicates a higher level of decodability for the token location probed. This effect remains consistent when removing verbalizations of the tokens in the target vocabulary, or using a probing task which does not rely on labels verbalized by the model (i.e. extracted directly from the environment). Although the loudest layer was not always the one most likely to be most decodable, it did lay on the plateau of highest decodability. Given a fitted J-Lens, loudness therefore offers a cheap first step before, or in place of, an exhaustive sweep. 

Following ICLR 2027 reviews, we recognize a limitation in the amount of evidence presented, and aim to address this in future research by testing J-Lens loudness on additional signals, tasks, and models.

\newpage




\newpage
\bibliography{bibliography}

@article{gurnee2026verbalizable,
  title={Verbalizable representations form a global workspace in language models},
  author={Gurnee, Wes and Sofroniew, Nicholas and Pearce, Adam and Piotrowski, Mateusz and Kauvar, Isaac and Chen, Runjin and Soligo, Anna and Bogdan, Paul and Ong, Euan and Wang, Rowan and others},
  journal={arXiv preprint arXiv:2607.15495},
  year={2026}
}

@article{belinkov2022probing,
  title={Probing classifiers: Promises, shortcomings, and advances},
  author={Belinkov, Yonatan},
  journal={Computational Linguistics},
  volume={48},
  number={1},
  pages={207--219},
  year={2022}
}

@article{lugoloobi2026llms,
  title={LLMs encode their failures: Predicting success from pre-generation activations},
  author={Lugoloobi, William and Foster, Thomas and Bankes, William and Russell, Chris},
  journal={arXiv preprint arXiv:2602.09924},
  year={2026}
}

@article{arghal2026behavioural,
  title={A behavioural and representational evaluation of goal-directedness in language model agents},
  author={Arghal, Raghu and Chen, Fade and Dalton, Niall and Kortukov, Evgenii and McNamara, Calum and Nalmpantis, Angelos and Nirvaan, Moksh and Sarti, Gabriele and Giulianelli, Mario},
  journal={arXiv preprint arXiv:2602.08964},
  year={2026}
}

@article{tigges2023linear,
  title={Linear representations of sentiment in large language models},
  author={Tigges, Curt and Hollinsworth, Oskar John and Geiger, Atticus and Nanda, Neel},
  journal={arXiv preprint arXiv:2310.15154},
  year={2023}
}

@inproceedings{ravichander2021probing,
  title={Probing the probing paradigm: Does probing accuracy entail task relevance?},
  author={Ravichander, Abhilasha and Belinkov, Yonatan and Hovy, Eduard},
  booktitle={Proceedings of the 16th Conference of the European Chapter of the Association for Computational Linguistics: Main Volume},
  pages={3363--3377},
  year={2021}
}

@online{nostalgebraist2020logitlens,
  author = {nostalgebraist},
  title  = {interpreting GPT: the logit lens},
  year   = {2020},
  month  = {8},
  url    = {https://alignmentforum.org},
  note   = {AI Alignment Forum}
}

@misc{mckenzie2026detectinghighstakesinteractionsactivation,
      title={Detecting High-Stakes Interactions with Activation Probes}, 
      author={Alex McKenzie and Urja Pawar and Phil Blandfort and William Bankes and David Krueger and Ekdeep Singh Lubana and Dmitrii Krasheninnikov},
      year={2026},
      eprint={2506.10805},
      archivePrefix={arXiv},
      primaryClass={cs.LG},
      url={https://arxiv.org/abs/2506.10805}, 
}

@article{kramar2026building,
  title={Building production-ready probes for gemini},
  author={Kram{\'a}r, J{\'a}nos and Engels, Joshua and Wang, Zheng and Chughtai, Bilal and Shah, Rohin and Nanda, Neel and Conmy, Arthur},
  journal={arXiv preprint arXiv:2601.11516},
  year={2026}
}

@inproceedings{pimentel2020information,
  title={Information-theoretic probing for linguistic structure},
  author={Pimentel, Tiago and Valvoda, Josef and Maudslay, Rowan Hall and Zmigrod, Ran and Williams, Adina and Cotterell, Ryan},
  booktitle={Proceedings of the 58th Annual Meeting of the Association for Computational Linguistics},
  pages={4609--4622},
  year={2020}
}

@inproceedings{
marks2024geometry,
title={The Geometry of Truth: Emergent Linear Structure in Large Language Model Representations of True/False Datasets},
author={Samuel Marks and Max Tegmark},
booktitle={First Conference on Language Modeling},
year={2024},
url={https://openreview.net/forum?id=aajyHYjjsk}
}

@misc{openai2025gptoss,
  title         = {GPT-OSS-120b \& GPT-OSS-20b model card},
  author        = {{OpenAI}},
  year          = {2025},
  eprint        = {2508.10925},
  archivePrefix = {arXiv},
  primaryClass  = {cs.CL},
  url           = {https://arxiv.org/abs/2508.10925}
}

@misc{qwen36_35b_a3b,
    title = {{Qwen3.6-35B-A3B}: Agentic Coding Power, Now Open to All},
    url = {https://qwen.ai/blog?id=qwen3.6-35b-a3b},
    author = {{Qwen Team}},
    month = {April},
    year = {2026}
}

@article{
sharkey2025open,
title={Open Problems in Mechanistic Interpretability},
author={Lee Sharkey and Bilal Chughtai and Joshua Batson and Jack Lindsey and Jeffrey Wu and Lucius Bushnaq and Nicholas Goldowsky-Dill and Stefan Heimersheim and Alejandro Ortega and Joseph Isaac Bloom and Stella Biderman and Adri{\`a} Garriga-Alonso and Arthur Conmy and Neel Nanda and Jessica Mary Rumbelow and Martin Wattenberg and Nandi Schoots and Joseph Miller and William Saunders and Eric J Michaud and Stephen Casper and Max Tegmark and David Bau and Eric Todd and Atticus Geiger and Mor Geva and Jesse Hoogland and Daniel Murfet and Thomas McGrath},
journal={Transactions on Machine Learning Research},
issn={2835-8856},
year={2025},
url={https://openreview.net/forum?id=91H76m9Z94},
note={Survey Certification}
}

@inproceedings{
goldowsky2025detecting,
title={Detecting Strategic Deception with Linear Probes},
author={Nicholas Goldowsky-Dill and Bilal Chughtai and Stefan Heimersheim and Marius Hobbhahn},
booktitle={Forty-second International Conference on Machine Learning},
year={2025},
url={https://openreview.net/forum?id=C5Jj3QKQav}
}

@article{alain2016understanding,
  title={Understanding intermediate layers using linear classifier probes},
  author={Alain, Guillaume and Bengio, Yoshua},
  journal={arXiv preprint arXiv:1610.01644},
  year={2016}
}

@article{skean2025layer,
  title={Layer by layer: Uncovering hidden representations in language models},
  author={Skean, Oscar and Arefin, Md Rifat and Zhao, Dan and Patel, Niket and Naghiyev, Jalal and LeCun, Yann and Shwartz-Ziv, Ravid},
  journal={arXiv preprint arXiv:2502.02013},
  year={2025}
}

@inproceedings{orgad2025llms,
  title={Llms know more than they show: On the intrinsic representation of llm hallucinations},
  author={Orgad, Hadas and Toker, Michael and Gekhman, Zorik and Reichart, Roi and Szpektor, Idan and Kotek, Hadas and Belinkov, Yonatan},
  booktitle={International Conference on Learning Representations},
  volume={2025},
  pages={66880--66913},
  year={2025}
}

@inproceedings{hewitt2019designing,
  title={Designing and interpreting probes with control tasks},
  author={Hewitt, John and Liang, Percy},
  booktitle={Proceedings of the 2019 conference on empirical methods in natural language processing and the 9th international joint conference on natural language processing (emnlp-ijcnlp)},
  pages={2733--2743},
  year={2019}
}

@inproceedings{giulianelli2018under,
  title={Under the hood: Using diagnostic classifiers to investigate and improve how language models track agreement information},
  author={Giulianelli, Mario and Harding, Jack and Mohnert, Florian and Hupkes, Dieuwke and Zuidema, Willem},
  booktitle={Proceedings of the 2018 EMNLP Workshop BlackboxNLP: Analyzing and Interpreting Neural Networks for NLP},
  pages={240--248},
  year={2018}
}

@inproceedings{gurnee2024language,
  title={Language models represent space and time},
  author={Gurnee, Wes and Tegmark, Max},
  booktitle={International Conference on Learning Representations},
  volume={2024},
  pages={2483--2503},
  year={2024}
}

@article{elazar2021amnesic,
  title={Amnesic probing: Behavioral explanation with amnesic counterfactuals},
  author={Elazar, Yanai and Ravfogel, Shauli and Jacovi, Alon and Goldberg, Yoav},
  journal={Transactions of the Association for Computational Linguistics},
  volume={9},
  pages={160--175},
  year={2021},
  publisher={MIT Press One Rogers Street, Cambridge, MA 02142-1209, USA journals-info~…}
}
\bibliographystyle{iclr2027_conference}
\newpage
\FloatBarrier
\appendix

\section{Capturing Signals}\label{appx: capturing_signals}

We construct $V_d$ and $V_g$ separately for each model using three steps.

First, we manually specify a small collection of \emph{anchor words} intended to describe the relevant signal. The anchors need not themselves correspond to individual tokens in the model vocabulary, although in our experiments they do.

Second, we perform a lexical expansion of the anchor set. For every vocabulary token, we compare its decoded string against each anchor using \texttt{difflib} and retain lexically similar tokens according to our selection criterion. For example, using \texttt{right} as an anchor for the direction signal recovers token strings such as \texttt{.right} and \texttt{/right}.

Third, we perform a semantic expansion, comparing the meaning of each vocabulary token against the same anchors using a semantic similarity model. This captures related tokens that need not be lexically similar to the anchors; for example, an anchor such as \texttt{right} can recover translations such as \texttt{derecho}.

We take the union of the lexical and semantic expansions to form $V_s$ and prune some of the words we deem to be false positives. This procedure is performed separately for each signal and model, since the two models use different token vocabularies. The resulting vocabulary sizes are found below in Table \ref{tab:vocabulary_tokens}.

\begin{table}[ht]
\centering
\caption{Subset sizes for each signal}
\label{tab:vocabulary_tokens}
\begin{tabularx}{\linewidth}{Xr}
\toprule
\textbf{Vocabulary} & \textbf{Number of tokens} \\
\midrule
GPT-OSS-20B direction & 446 \\
GPT-OSS-20B grid & 642 \\
Qwen3.6-35B-A3B direction & 643 \\
Qwen3.6-35B-A3B grid & 533 \\
\bottomrule
\end{tabularx}
\end{table}
\FloatBarrier
\section{Experimental details}
\label{appx:details}

\paragraph{Data.}
Each example is the reasoning chain of a single environment step, and training and evaluation sets are split by chain. The training sets contain $2{,}880$ chains for GPT-OSS-20B and $556$ for Qwen3.6-35B-A3B. The held-out evaluation sets contain $720$ chains for GPT-OSS-20B and $52$ for Qwen3.6-35B-A3B, giving $14{,}391$ and $3{,}120$ positions per top-$N$ subset, respectively, and $14{,}472$ positions per sentence-level subset for GPT-OSS-20B. For Qwen3.6-35B-A3B, $20\%$ of reasoning-token positions are subsampled uniformly before any selection.

\paragraph{J-Lens.}
For each model we estimate $\bar J_l$ on the model's own grid trajectories, drawn from the training split only and balanced over grid size and obstacle complexity. Each context is the exact prompt the model saw followed by its full completion. For a context, row $i$ of $J_l$ is the gradient of $\sum_{p'} h_T[p', i]$ with respect to $h_l[p]$, averaged over source positions $p$ in a fit window. $\bar J_l$ is the mean of these over contexts, with $T$ the final block (where the lens is the identity).
 
For GPT-OSS-20B we use $2{,}880$ contexts each truncated to 1024 tokens and exclude the 414-token instruction prefix shared by every prompt from the training, so the averaged positions cover the grid and the opening of the reasoning. We stop early once the mean relative change of $\bar J_l$ falls below 0.2\%.
 
For Qwen3.6-35B-A3B, whose reasoning chains have a median of about 18k tokens, we use 144 contexts (4 per size $\times$ complexity cell) and fit the nine full-attention layers (3, 7, \dots, 35) against layer 39. Positions are a $512$-token window ending halfway through the reasoning chain. The preceding tokens are run through the model without gradient tracking, so the window activations keep their full context at a fraction of the cost.
 In both cases the transported state is passed through the model's own final RMSNorm before the unembedding, $\mathrm{logits} = W_U\,\mathrm{RMSNorm}(\bar J_l h_l)$, using the zero-centred $(1+w)$ scale for Qwen3.6-35B-A3B.

\paragraph{Probes.}

Probes are trained on the residual stream at a single layer, after per-dimension standardisation using training-set statistics. The linear probe is a multinomial logistic regression. The MLP has one hidden layer of 1024 ReLU units, with dropout 0.1 for the Qwen3.6-35B-A3B direction and goal-side probes and none for the GPT-OSS-20B probes and the grid probes. Both are trained with AdamW (learning rate $3\times10^{-4}$, weight decay $10^{-3}$) for 50 epochs, using a class-balanced cross-entropy loss (weights $n/(K n_c)$), and we report the final-epoch model without early stopping. Train and validation sets are split by trajectory.

The direction probe predicts one of the four actions from a single reasoning-token activation. Its label is the model's local belief: the action it outputs when its reasoning is truncated at that token, the reasoning channel is closed and the final-answer channel is opened.

The goal-side probe is a binary probe on the same positions, chains and split as the direction probe. Its label compares the goal's column with the agent's column in the environment's grid parameters (\texttt{LEFT} if smaller, \texttt{RIGHT} if larger); steps where they are equal are dropped. This keeps $2{,}521$ of $2{,}880$ training and $628$ of $720$ held-out chains for GPT-OSS-20B, and $40$ of $52$ held-out chains for Qwen3.6-35B-A3B ($9$ of them \texttt{LEFT}). As the label is constant within a chain, the effective size of the Qwen3.6-35B-A3B evaluation set is small.

For the layer sweeps, we re-extract the residual stream at every layer (GPT-OSS-20B: $3$--$22$; Qwen3.6-35B-A3B: $3$--$38$) at exactly the positions used at $l^*_d$, and train probes with the hyperparameters above. Re-extracted activations at $l^*_d$ match the original ones up to bf16 noise, and the re-trained probes at $l^*_d$ reproduce the original accuracies to within $0.006$. We report the final epoch, averaged over $5$ (GPT-OSS-20B) or $3$ (Qwen3.6-35B-A3B) seeds.

The J-Lens classifier maps each activation through $\bar J_l$, the final RMSNorm and the unembedding, and predicts the action token with the highest log-probability. It uses the same held-out positions and labels as the direction probes, and needs no training.

The grid probe predicts the contents of a queried cell (agent, wall, goal, empty, or padding). The cell's (row, column) coordinates are appended to the activation as two unnormalised scalar features. Grids are padded to $15\times15$, and padding is kept as its own class. We draw 25 cells uniformly per step with a fixed per-step seed, so every token-selection arm is evaluated on the same cells.

\paragraph{Decile analysis.}
Deciles are computed over all held-out tokens pooled across chains. Confidence intervals are $95\%$ intervals from $300$ bootstrap resamples of held-out chains. For the signal-word exclusion, every token of $V_s$ and the two tokens on either side within the same chain are removed before recomputing deciles.

\paragraph{Compute.}
We used a collection of GPUs to run our experiments. For activation gathering, we used primarily A100 GPUs, while all other computations were performed on cheaper GPUs such as RTX 2000 Ada and RTX PRO 4500.
Part of our system uses the MooseFS file system, which slows down navigation and path finding methods such as \texttt{glob} significantly. This contributed to a much slower compute time than necessary. In total, to reproduce the experiments in this paper using 1 GPU at a time, we estimate $90$ hours. Most of the compute is dominated by the need to fit our own J-Lens, something which we expect to no longer be an issue once more trained J-Lens weights are open sourced.

\textbf{Prompt.}
\begin{lstlisting}[style=prompt]
<|start|>system<|message|>You are ChatGPT, a large language model trained by OpenAI.
Knowledge cutoff: 2024-06
Current date: 2026-01-17

Reasoning: medium

# Valid channels: analysis, commentary, final. Channel must be included for every message.<|end|><|start|>user<|message|># Instructions

You are controlling an agent in a grid-based environment with full observability. The agent can move in four directions: up, down, left, and right. The environment contains walls, open spaces, and a goal location. The following symbols are used in the grid representation:

Legend:
---------------
#: Wall
_: Open Space (can be visited)
G: Goal
A: Current agent position
----------------

Your objective is to navigate from the current position (A) to the goal (G) while avoiding walls (#). Importantly, the agent should aim to reach the goal using the least amount of steps possible. You must decide your next move based on the provided information.

You will receive the current state of the grid as a NxM matrix of symbols separated by whitespaces, and with coordinates for each row an column. For example, given the 4x5 grid:

  0 1 2 3 4
0 # # # # #
1 # _ _ G #
2 # A _ _ #
3 # # # # #

The goal is in position (1,3), while the agent's position A is (2,1).

The agent's possible actions at each step are:

Actions:
- UP: Move Up
- DOWN: Move Down
- LEFT: Move Left
- RIGHT: Move Right

Your final answer should be a valid JSON object provided in the following form:
{
  "action": "<UP|DOWN|LEFT|RIGHT>",
}

DO NOT INCLUDE ANY `json` or `jsonb` in your response NOR TICK MARKS LIKE THIS: ```json ```. Start with { and end with } exactly.

# Inputs

Current grid state:
{{grid_state}}

<|end|><|start|>assistant
\end{lstlisting}

For Qwen3.6-35B-A3B we replace the reasoning tokens with \texttt{<think>} and \texttt{</think>}.

\FloatBarrier
\section{Additional results}
\label{appx:additional_results}

\FloatBarrier
\subsection{Logit-Lens layer profiles}
\begin{figure}[h]
  \centering
  \begin{tabular}{@{}c@{\hspace{2pt}}cc@{}}
    & \small Direction & \small Grid \\
    \rotatebox{90}{\small\hspace{1.2em}GPT-OSS-20B} &
    \includegraphics[width=0.47\linewidth]{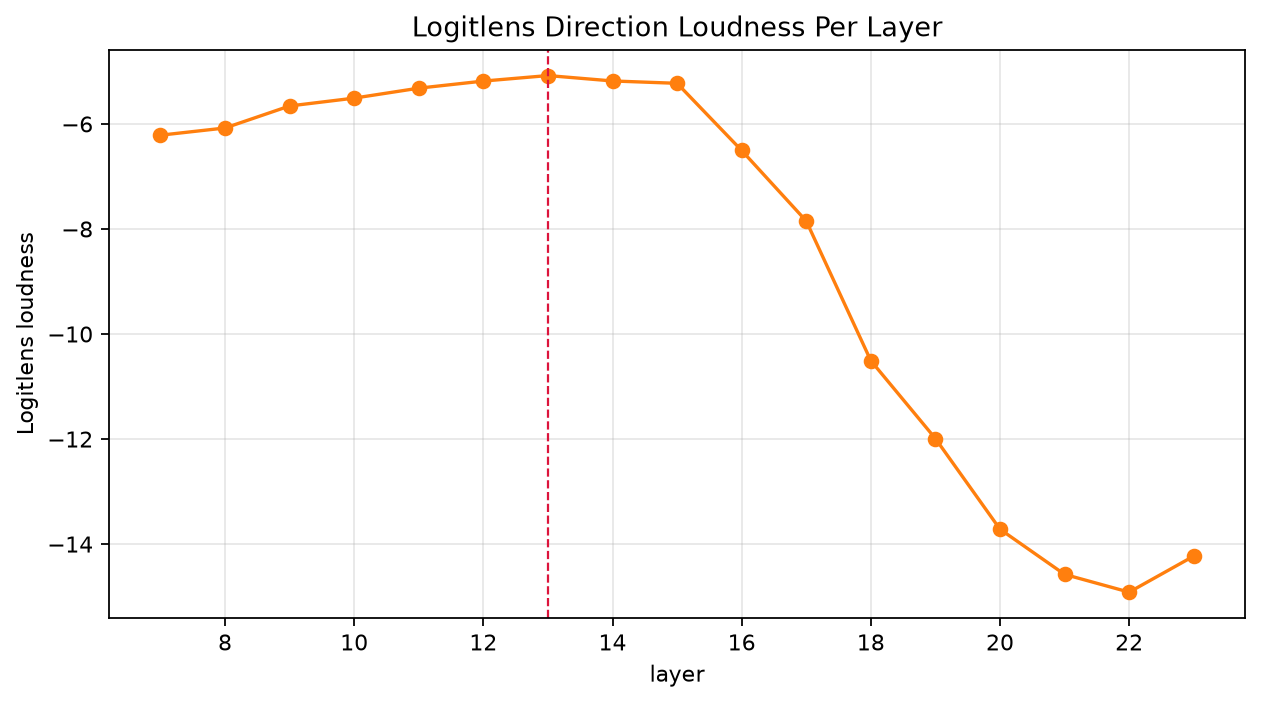} &
    \includegraphics[width=0.47\linewidth]{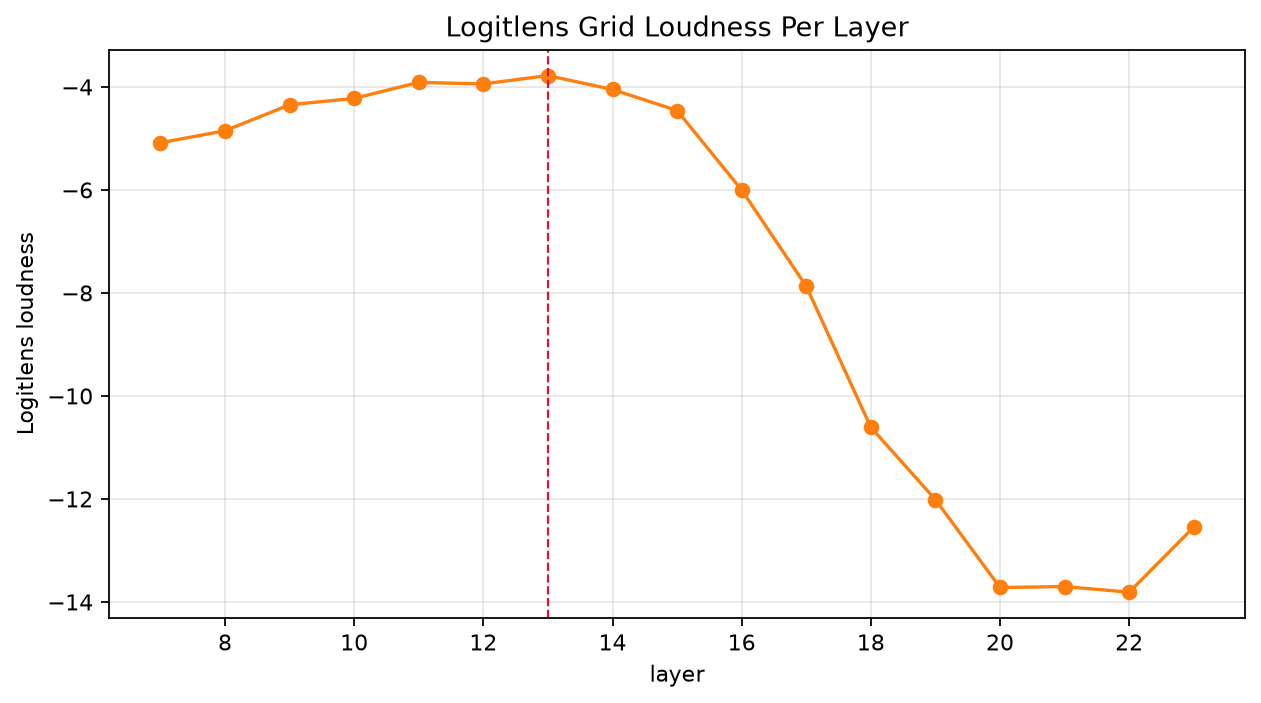} \\
    \rotatebox{90}{\small\hspace{0.8em}Qwen3.6-35B-A3B} &
    \includegraphics[width=0.47\linewidth]{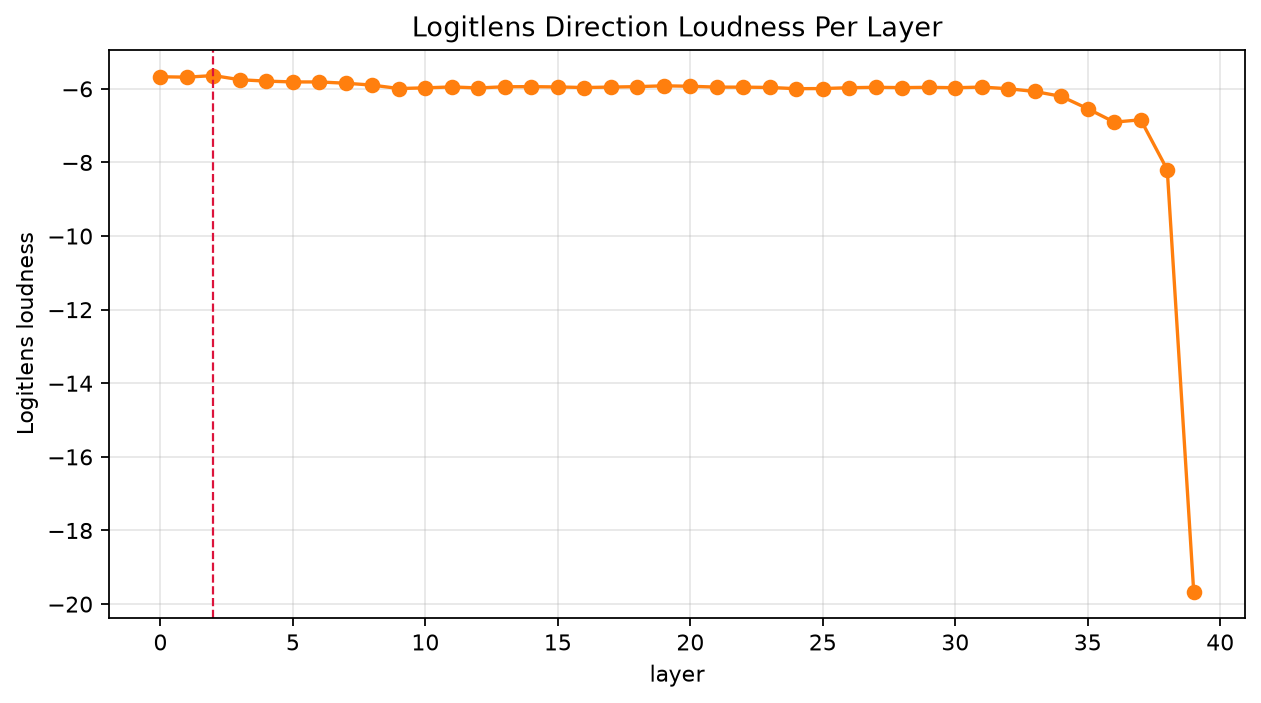} &
    \includegraphics[width=0.47\linewidth]{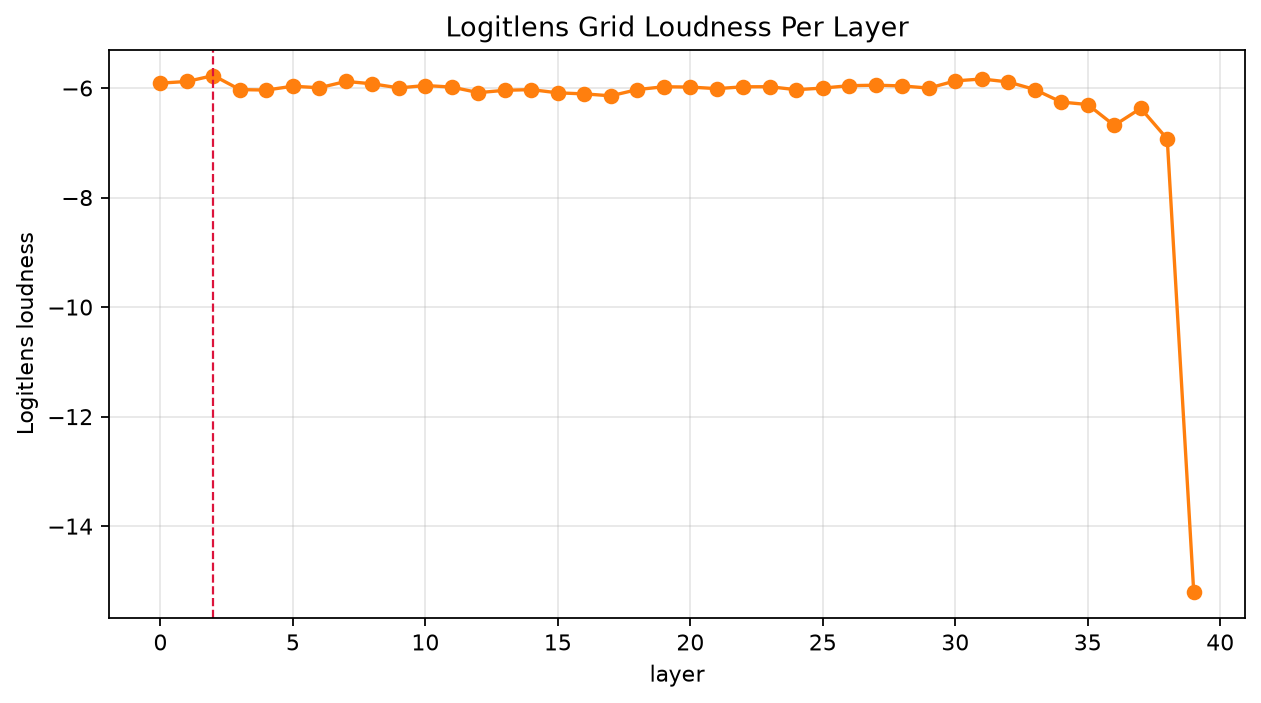} \\
  \end{tabular}
  \caption{Mean Logit-Lens loudness per layer. Dashed line: argmax.}
  \label{fig:appx_logitlens_layers}
\end{figure}

For GPT-OSS-20B, Logit-Lens loudness peaks at layer $13$ for both signals and declines earlier than J-Lens loudness (Figure~\ref{fig:layer_profiles}). For Qwen3.6-35B-A3B, the profile is flat from layer $0$ to $33$, so its argmax (layer $2$) carries no information about depth (Figure~\ref{fig:appx_logitlens_layers}).

\FloatBarrier
\subsection{Agreement between J-Lens and Logit-Lens loudness}
\begin{figure}[h]
  \centering
  \includegraphics[width=0.35\linewidth]{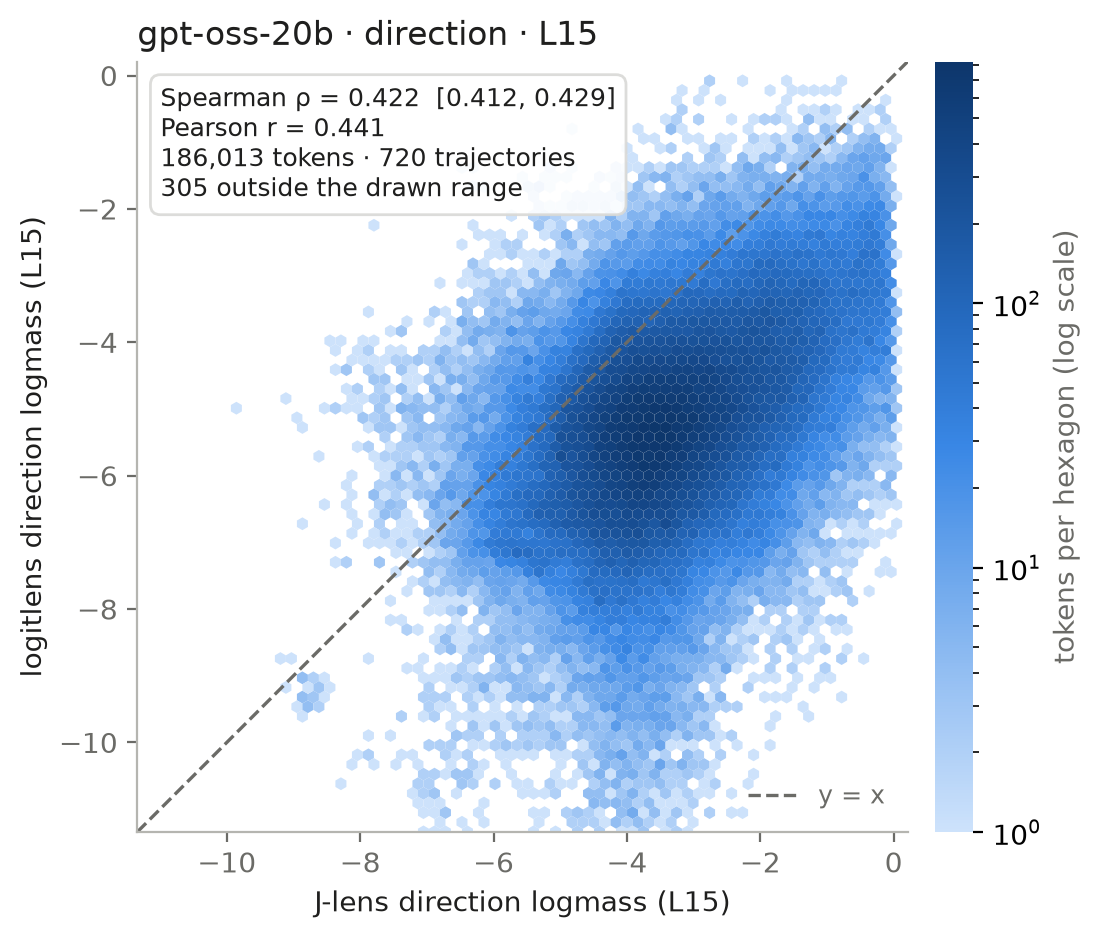}
  \includegraphics[width=0.35\linewidth]{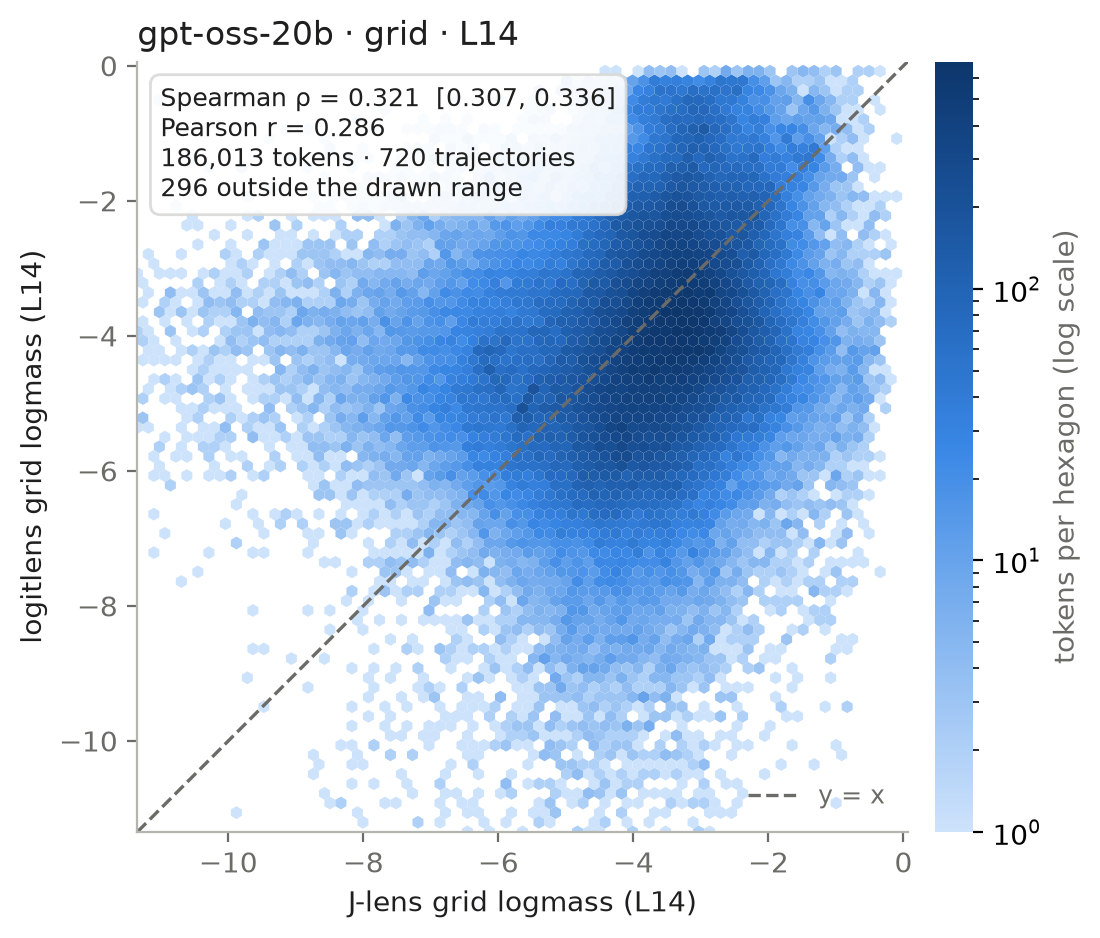} \\
  \includegraphics[width=0.35\linewidth]{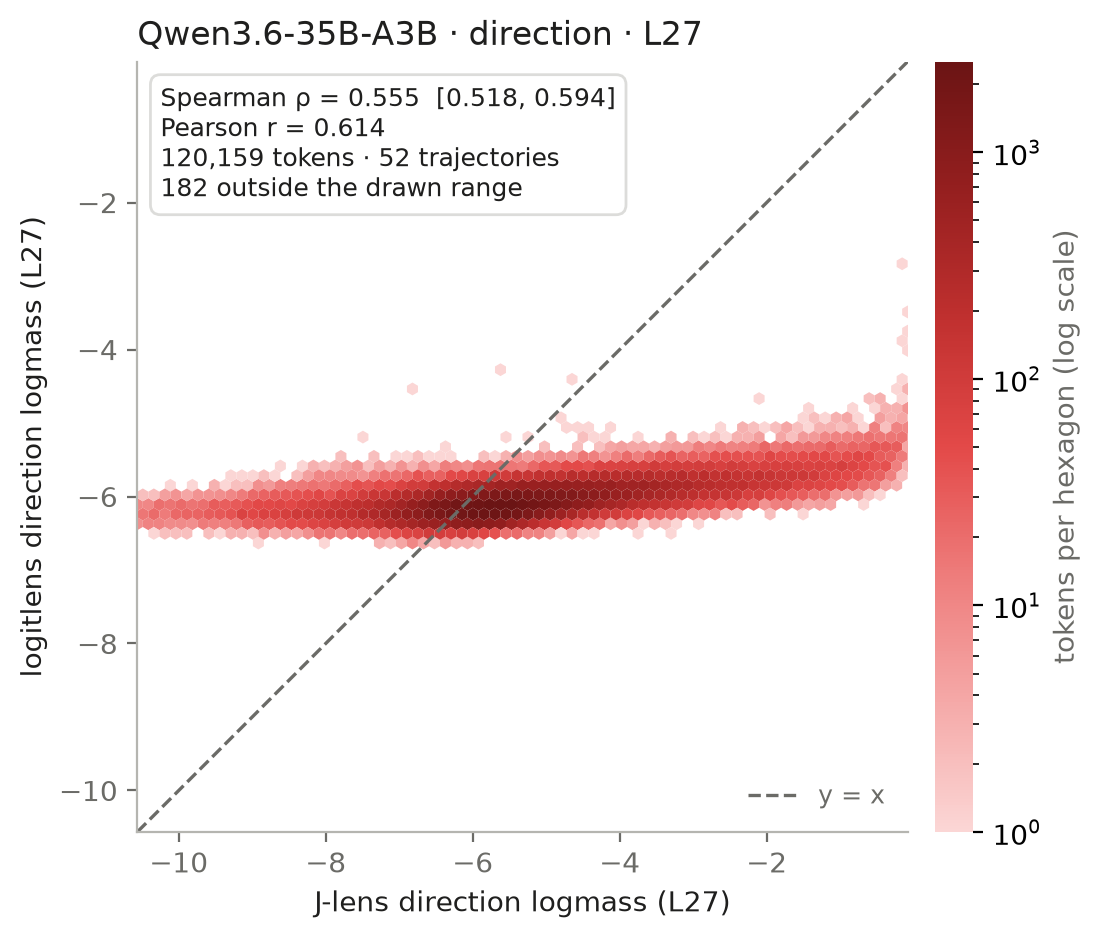}
  \includegraphics[width=0.35\linewidth]{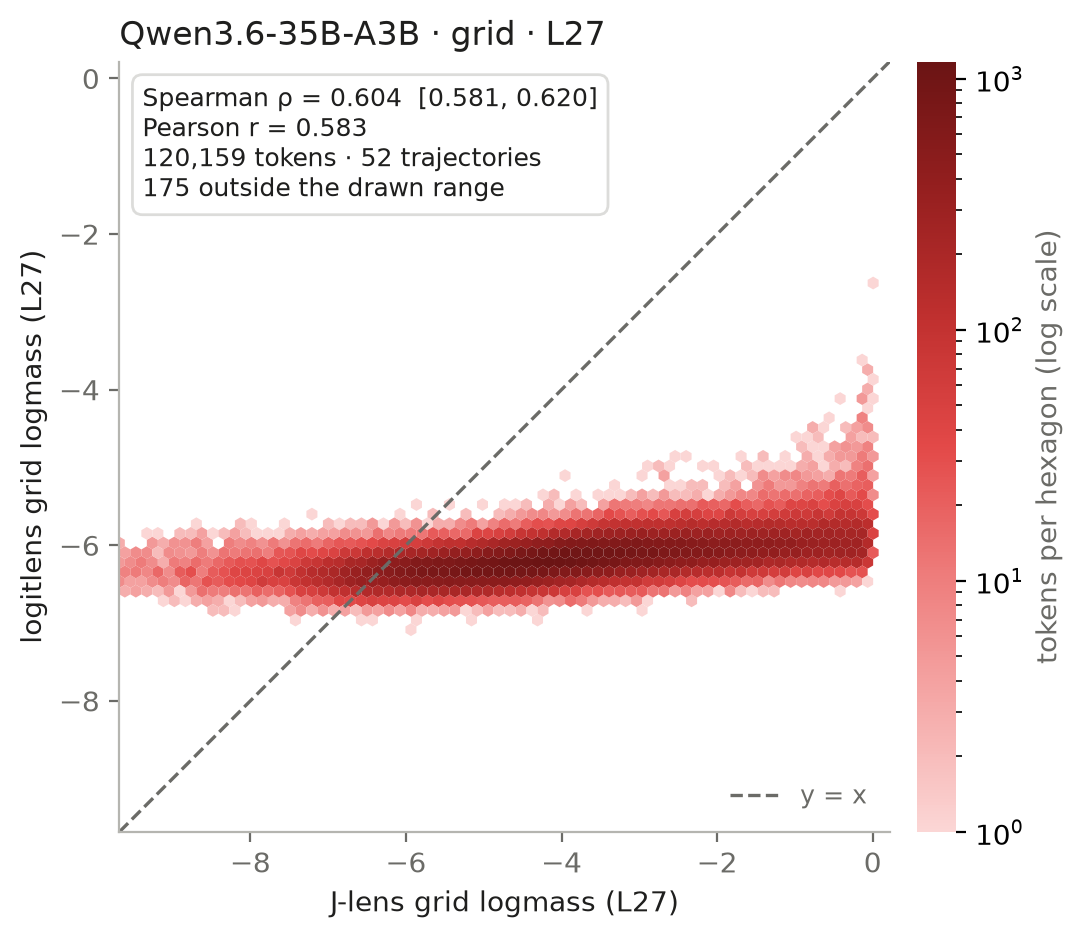}
  \caption{Token-level J-Lens vs.\ Logit-Lens loudness at $l^*_s$. Top: GPT-OSS-20B; bottom: Qwen3.6-35B-A3B; left: direction; right: grid.}
  \label{fig:appx_lens_vs_lens}
\end{figure}

At the token level, J-Lens and Logit-Lens loudness for the same signal are only moderately correlated (Spearman $\rho$ between $0.32$ and $0.60$; Figure~\ref{fig:appx_lens_vs_lens}), so the two lenses select different top-$N$ positions. For Qwen3.6-35B-A3B, Logit-Lens loudness varies within a narrow band, while J-Lens loudness spans roughly ten nats.

\FloatBarrier
\subsection{Agreement between direction and grid loudness}
\begin{figure}[h]
  \centering
  \includegraphics[width=0.35\linewidth]{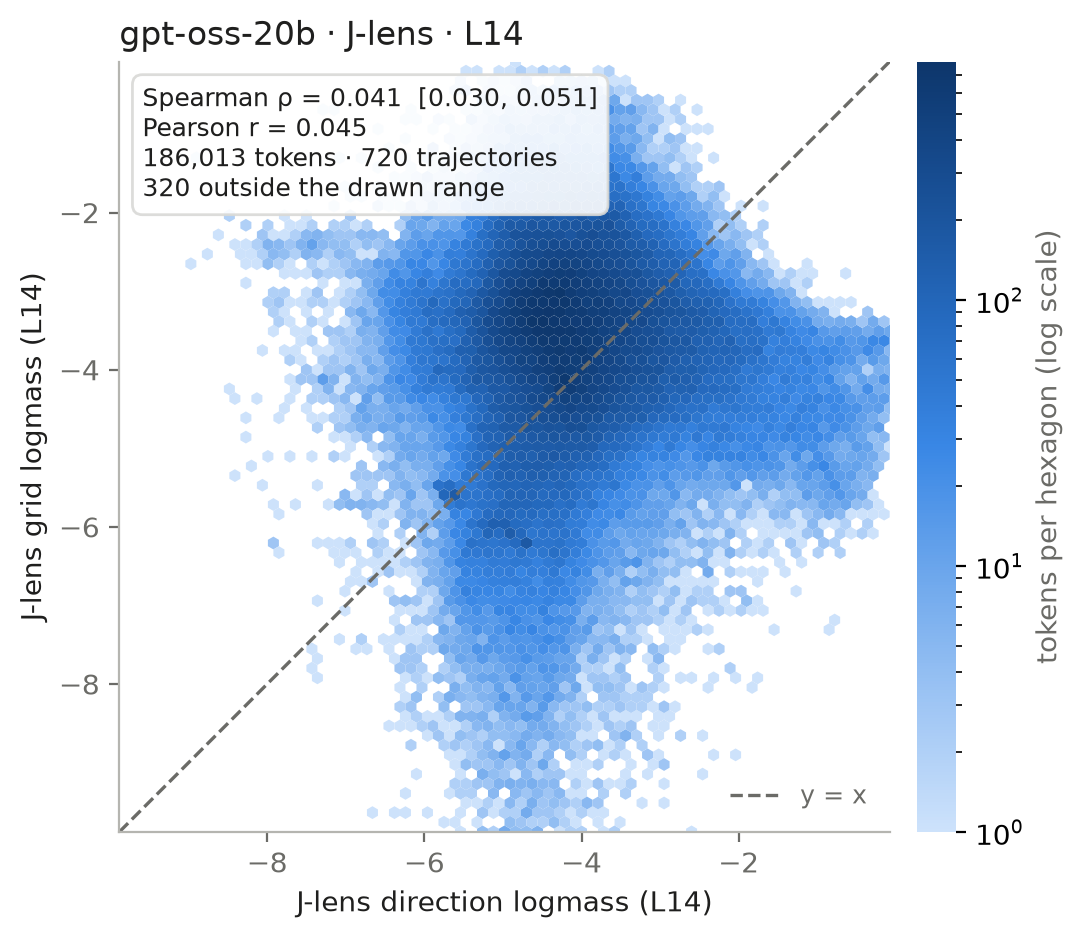}
  \includegraphics[width=0.35\linewidth]{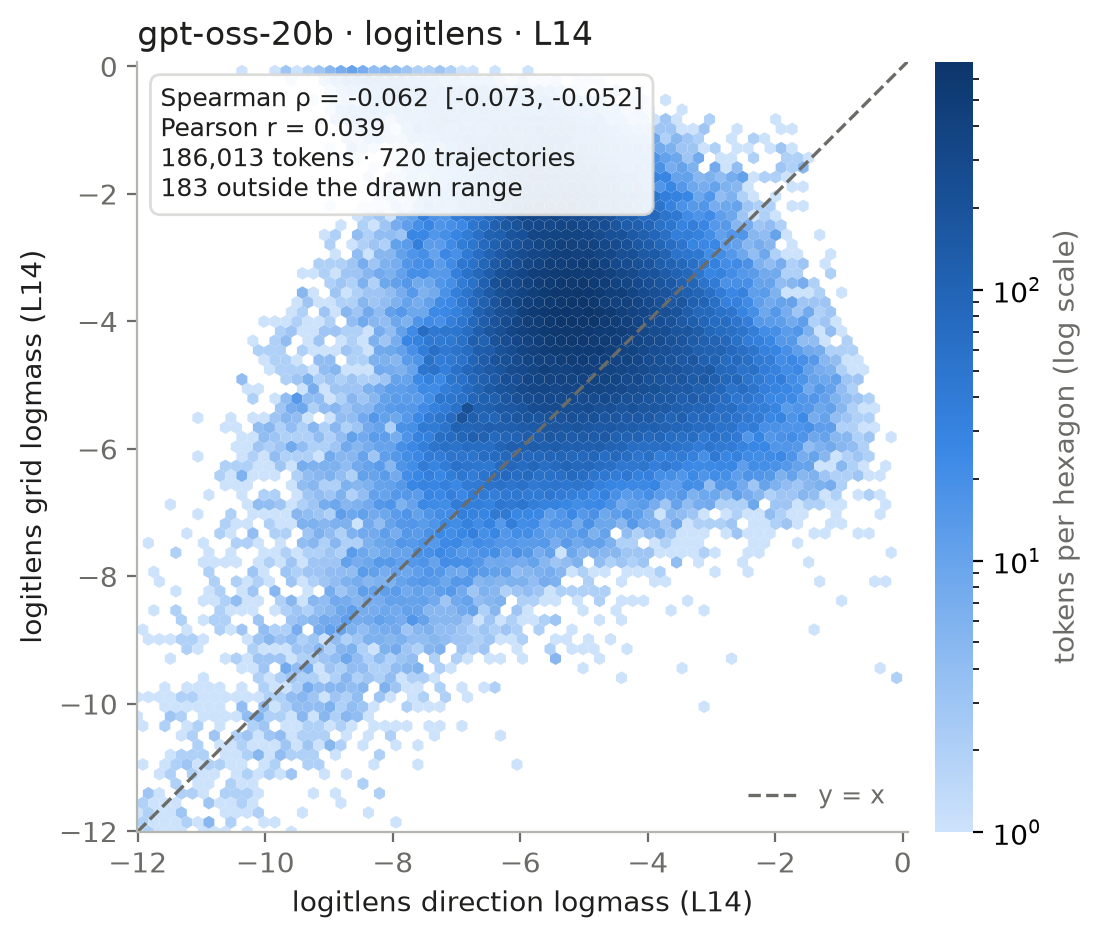} \\
  \includegraphics[width=0.35\linewidth]{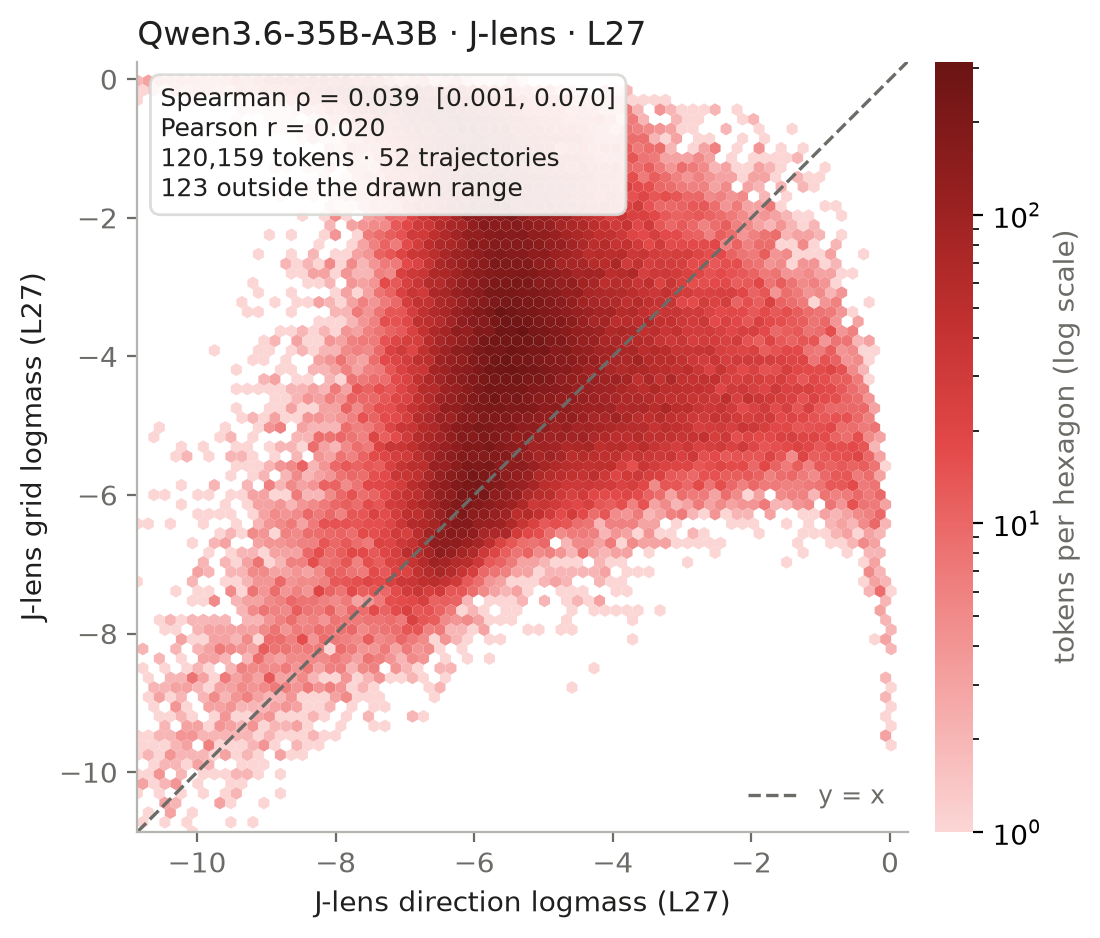}
  \includegraphics[width=0.35\linewidth]{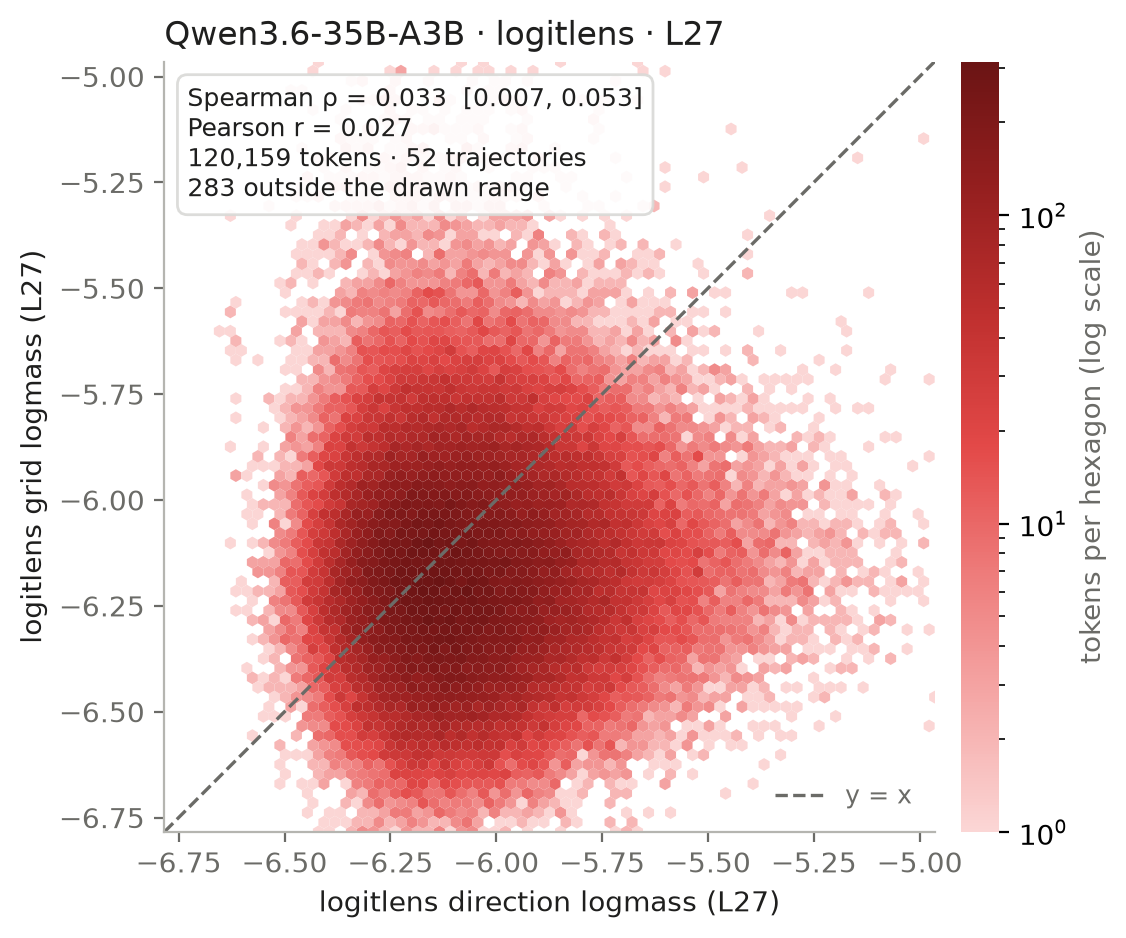}
  \caption{Token-level direction ($x$) vs.\ grid ($y$) loudness. Top: GPT-OSS-20B; bottom: Qwen3.6-35B-A3B; left: J-Lens; right: Logit Lens.}
  \label{fig:appx_signal_vs_signal}
\end{figure}

Direction and grid loudness are nearly uncorrelated at the token level under both lenses and in both models ($|\rho| \le 0.062$; Figure~\ref{fig:appx_signal_vs_signal}).

\FloatBarrier
\subsection{Grid cross-selection results}
\label{appx:full_tables}

Table~\ref{tab:appx_topn_grid} reports every combination of training criterion (rows) and evaluation criterion (columns) for top-$N$ selection at $l^*_s$ for the grid signal, for both probe architectures, on the same position subsets as the direction results in Table~\ref{tab:topn_dir}. Accuracy is balanced over the four cell classes, excluding padding. \textbf{Bold} marks the best evaluation subset for each probe (computed on unrounded values) and \underline{underlining} marks in-distribution evaluation. Balanced accuracy; chance is $0.25$.

\begin{table}[h]
  \centering
  \caption{Grid, top-$N$ selection. Rows: training criterion; columns: evaluation subset.}
  \label{tab:appx_topn_grid}
  \small
  \setlength{\tabcolsep}{4.5pt}
  \begin{tabular}{llcccccc}
    \toprule
    & & \multicolumn{3}{c}{GPT-OSS-20B} & \multicolumn{3}{c}{Qwen3.6-35B-A3B} \\
    \cmidrule(lr){3-5}\cmidrule(lr){6-8}
    Trained on & Probe & J-Lens & Logit Lens & Random & J-Lens & Logit Lens & Random \\
    \midrule
    J-Lens     & Linear & \underline{.292} & \textbf{.294} & .283 & \underline{.258} & .240 & \textbf{.263} \\
               & MLP    & \underline{.438} & \textbf{.445} & .402 & \underline{\textbf{.371}} & .346 & .349 \\
    Logit Lens & Linear & .289 & \underline{\textbf{.303}} & .286 & \textbf{.252} & \underline{.247} & .252 \\
               & MLP    & .421 & \underline{\textbf{.473}} & .403 & \textbf{.374} & \underline{.370} & .345 \\
    Random     & Linear & .290 & \textbf{.303} & \underline{.289} & \textbf{.245} & .234 & \underline{.243} \\
               & MLP    & .413 & \textbf{.432} & \underline{.410} & \textbf{.360} & .354 & \underline{.344} \\
    \bottomrule
  \end{tabular}
\end{table}

\FloatBarrier
\subsection{Probe accuracy by loudness decile}

\begin{figure}[h]
  \centering
  \begin{tabular}{@{}c@{\hspace{2pt}}cc@{}}
    & \small GPT-OSS-20B & \small Qwen3.6-35B-A3B \\
    \rotatebox{90}{\small\hspace{0.6em}Direction probes} &
    \includegraphics[width=0.45\linewidth]{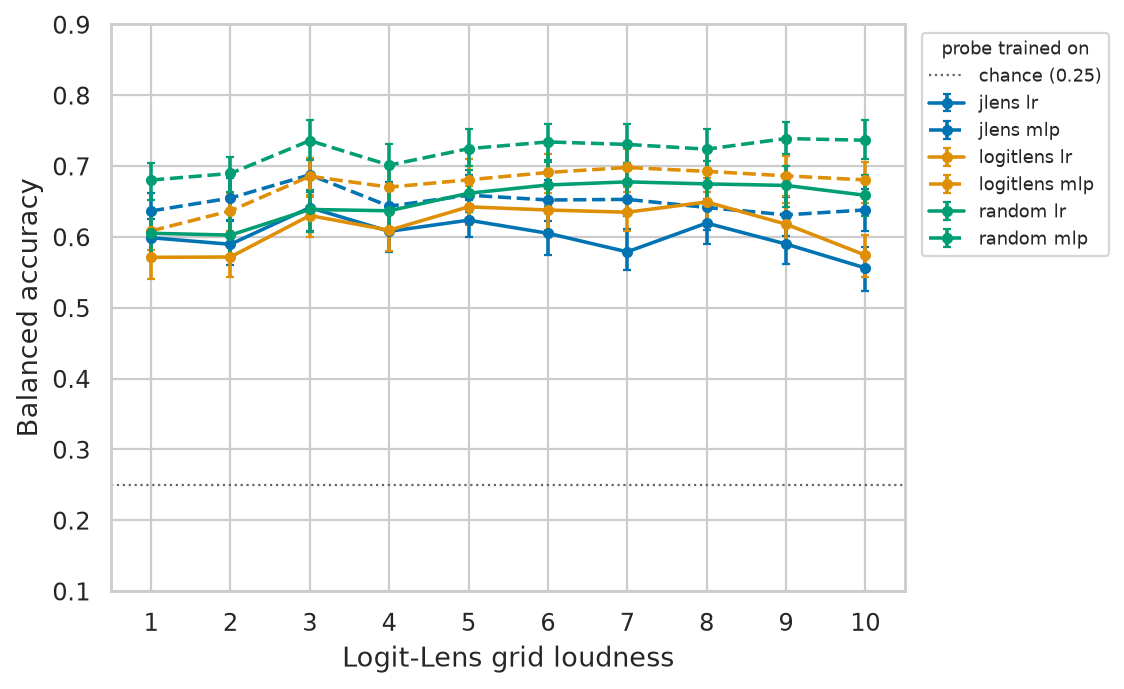} &
    \includegraphics[width=0.45\linewidth]{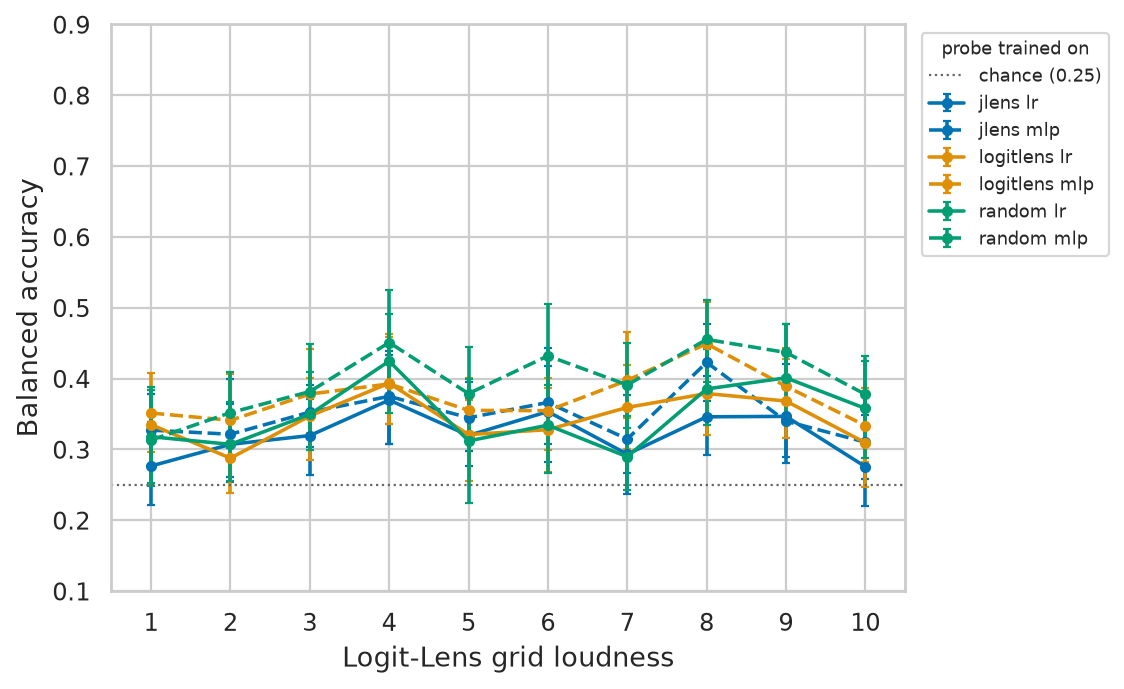} \\
    \rotatebox{90}{\small\hspace{1.4em}Grid probes} &
    \includegraphics[width=0.45\linewidth]{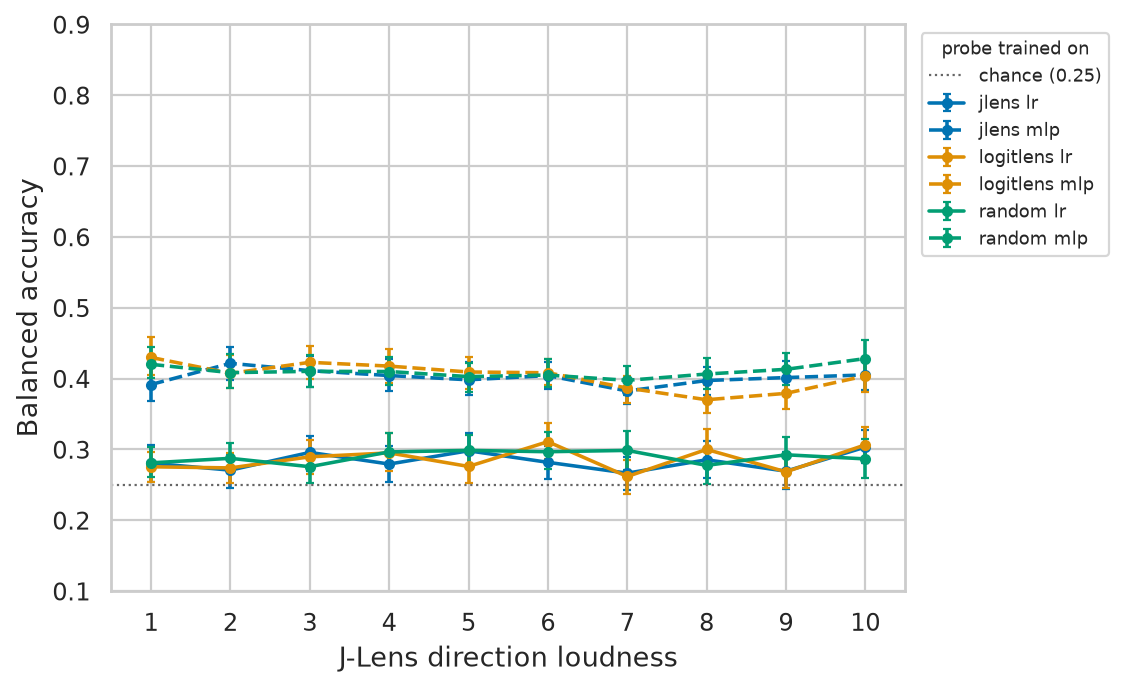} &
    \includegraphics[width=0.45\linewidth]{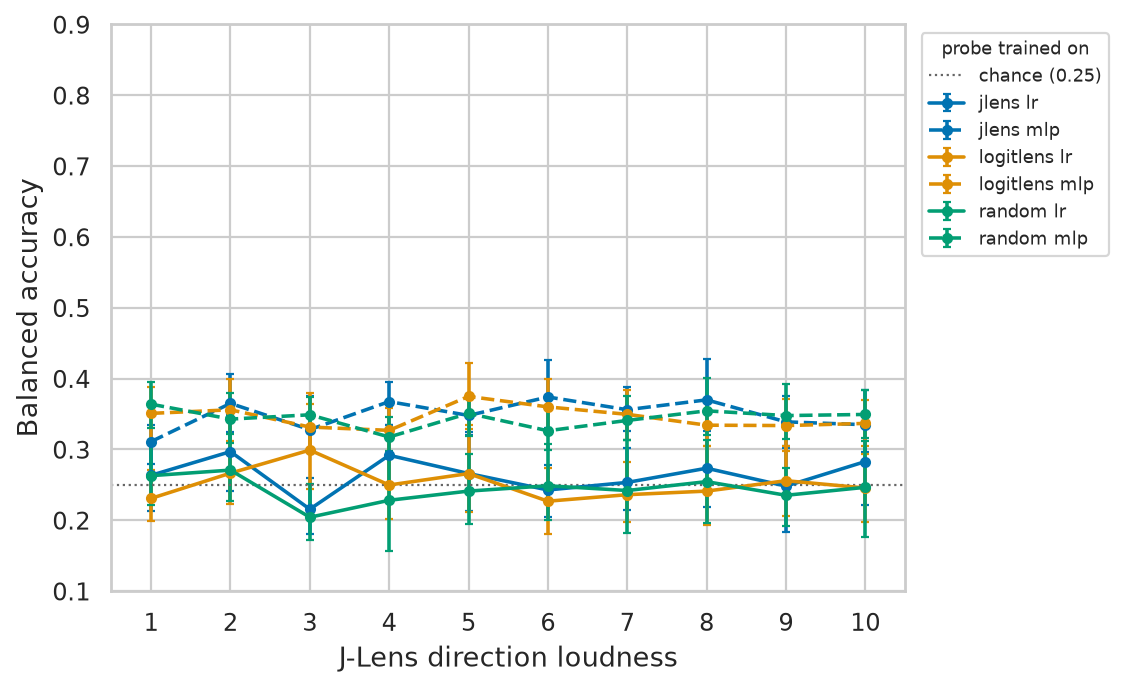} \\
  \end{tabular}
  \vspace{-6pt}
  \caption{Opposite-signal controls. Top: direction probes by Logit-Lens grid loudness. Bottom: grid probes by J-Lens direction loudness.}
  \label{fig:appx_decile_opposite}
    \vspace{-12pt}
\end{figure}

All panels follow the conventions of Figure~\ref{fig:deciles_direction}: balanced accuracy on held-out tokens, binned into deciles of loudness pooled across reasoning chains; solid lines are linear probes and dashed lines MLP probes. In the signal-word-excluded variants, every token of $V_s$ and the two tokens on either side are removed and deciles are recomputed on the remaining tokens. This removes $27.9\%$ (GPT-OSS-20B) and $10.0\%$ (Qwen3.6-35B-A3B) of tokens for direction, and $29.1\%$ and $34.0\%$ for grid. As opposite-signal controls, direction probes binned by Logit-Lens grid loudness (the J-Lens version is Figure~\ref{fig:selectivity}) and grid probes binned by direction loudness under either lens show no consistent trend (Figure~\ref{fig:appx_decile_opposite}).

\begin{figure}[h]
  \centering
  \begin{tabular}{@{}c@{\hspace{2pt}}cc@{}}
    & \small Ranked by J-Lens loudness & \small Ranked by Logit-Lens loudness \\
    \rotatebox{90}{\small\hspace{1.6em}GPT-OSS-20B} &
    \includegraphics[width=0.45\linewidth]{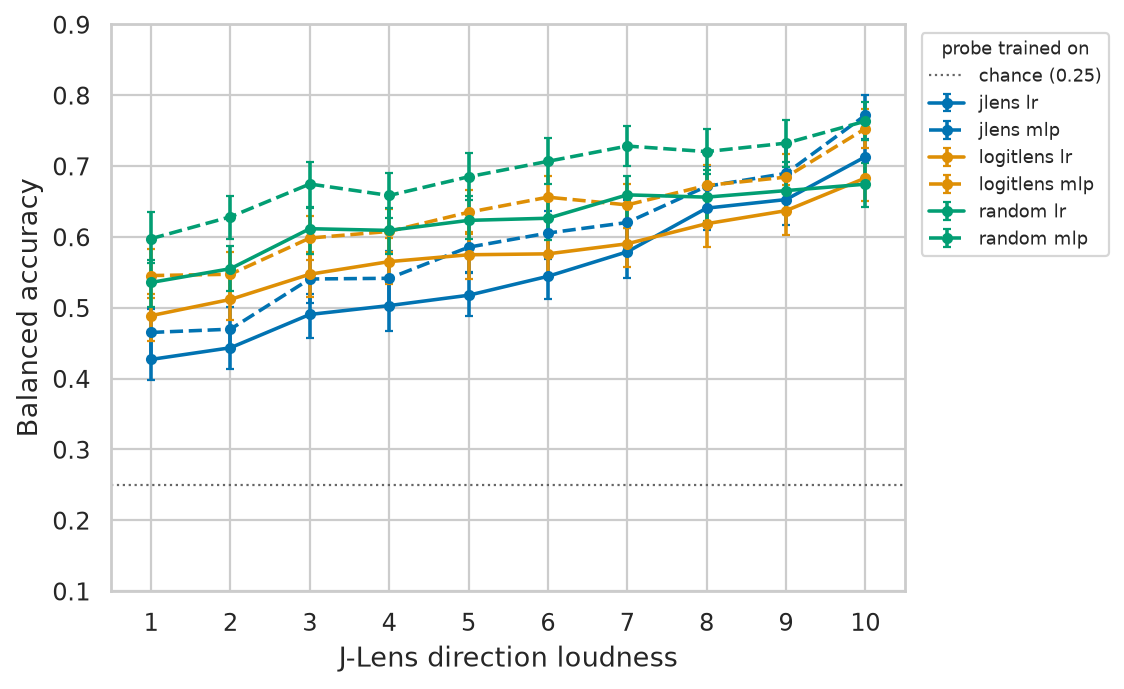} &
    \includegraphics[width=0.45\linewidth]{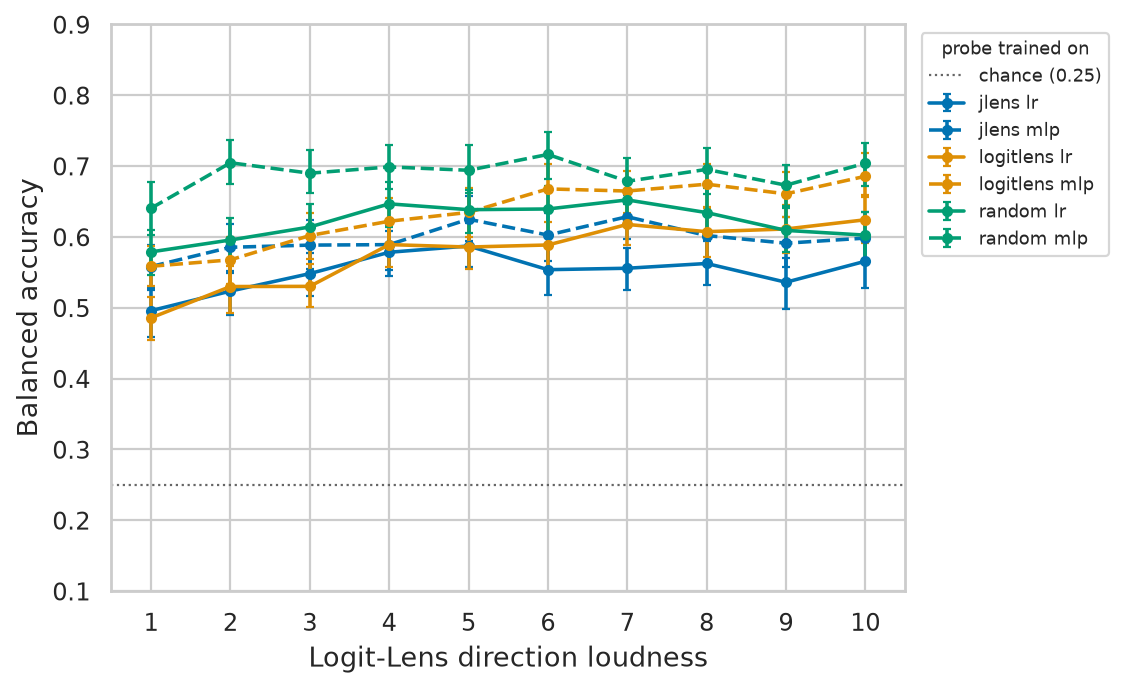} \\
    \rotatebox{90}{\small\hspace{1.2em}Qwen3.6-35B-A3B} &
    \includegraphics[width=0.45\linewidth]{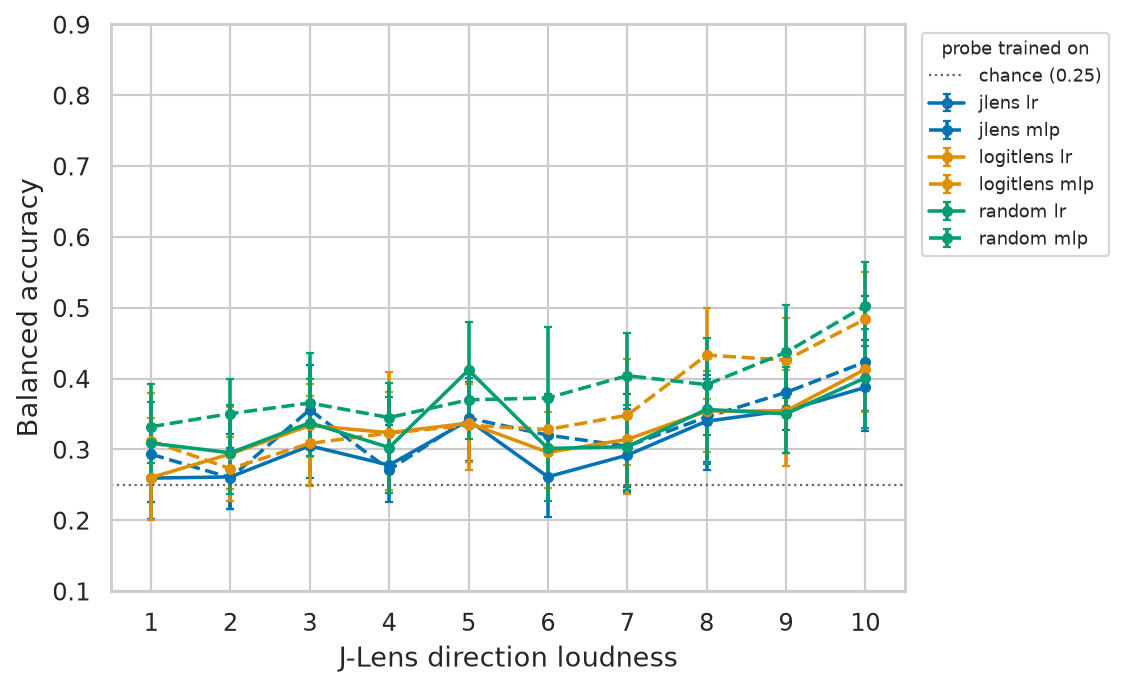} &
    \includegraphics[width=0.45\linewidth]{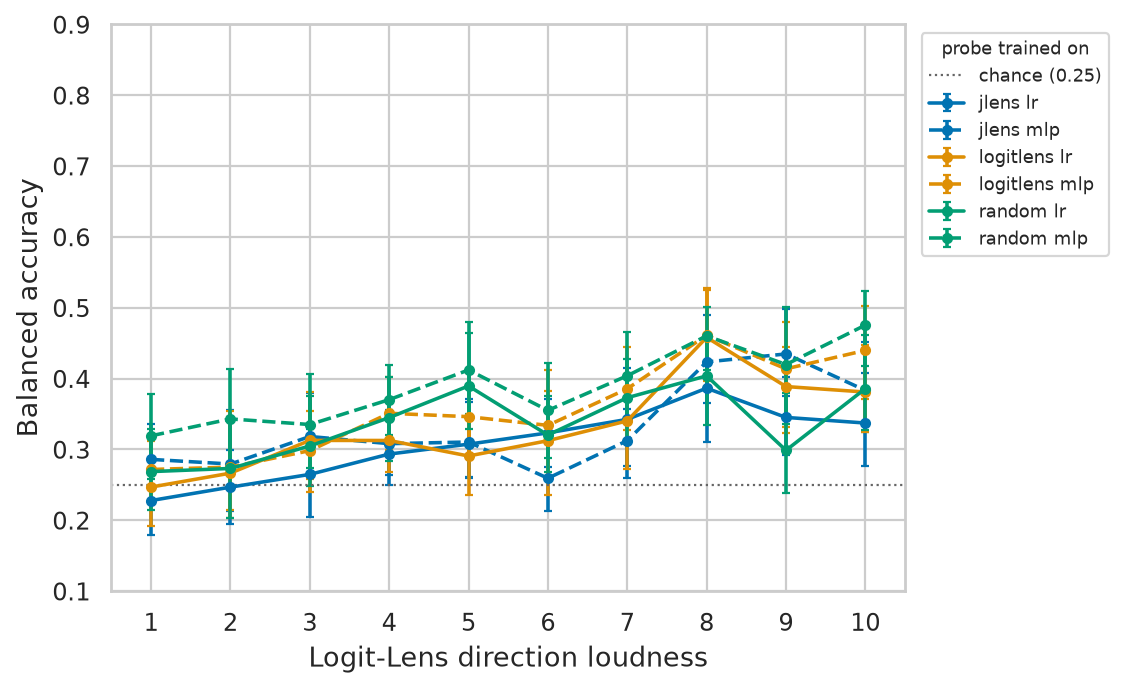} \\
  \end{tabular}
  \caption{Direction probes by direction-loudness decile, excluding direction words and their $\pm2$ neighbours.}
  \label{fig:appx_decile_nowords}
\end{figure}

\begin{figure}[h]
  \centering
  \begin{tabular}{@{}c@{\hspace{2pt}}cc@{}}
    & \small Ranked by J-Lens loudness & \small Ranked by Logit-Lens loudness \\
    \rotatebox{90}{\small\hspace{1.6em}GPT-OSS-20B} &
    \includegraphics[width=0.45\linewidth]{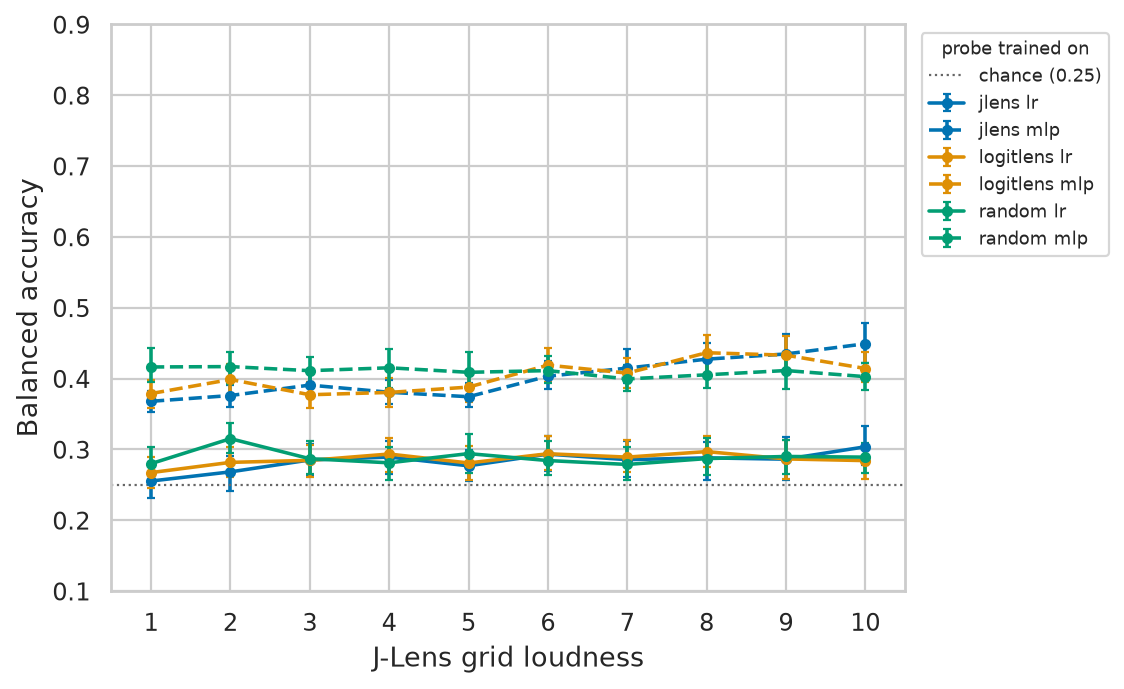} &
    \includegraphics[width=0.45\linewidth]{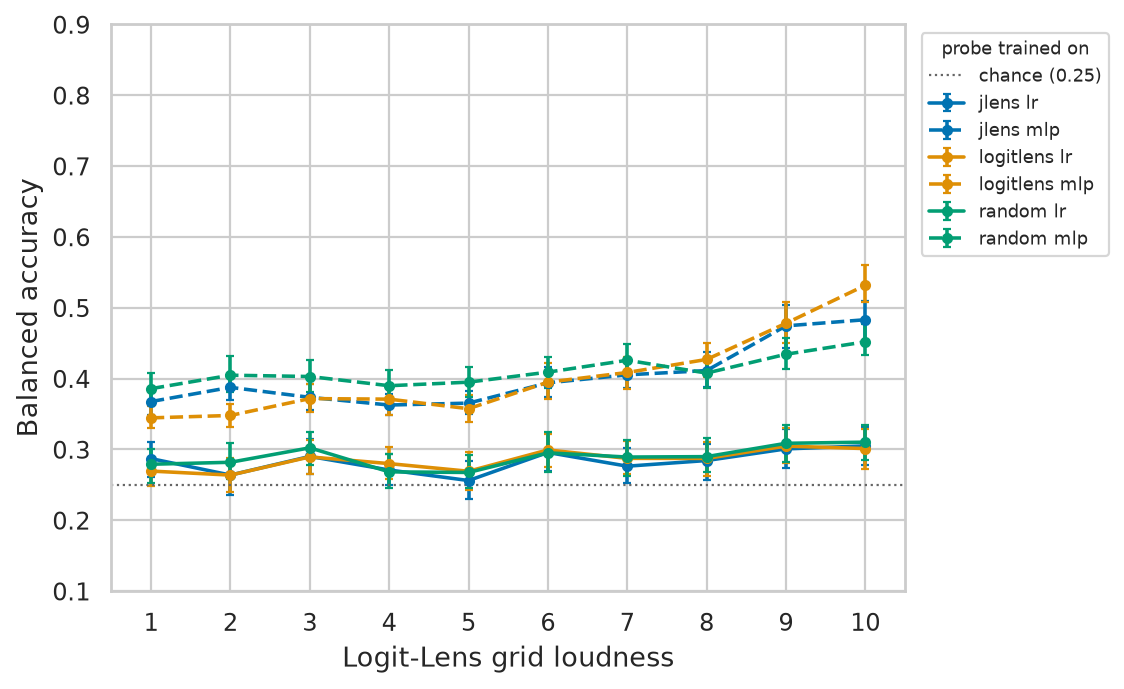} \\
    \rotatebox{90}{\small\hspace{1.2em}Qwen3.6-35B-A3B} &
    \includegraphics[width=0.45\linewidth]{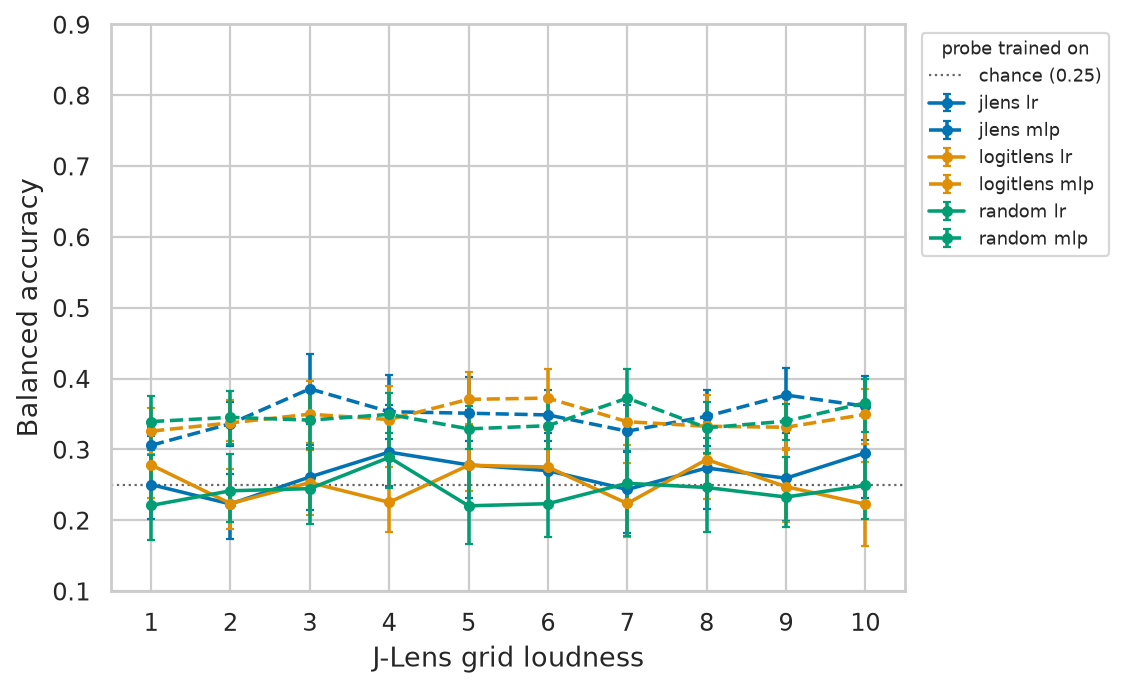} &
    \includegraphics[width=0.45\linewidth]{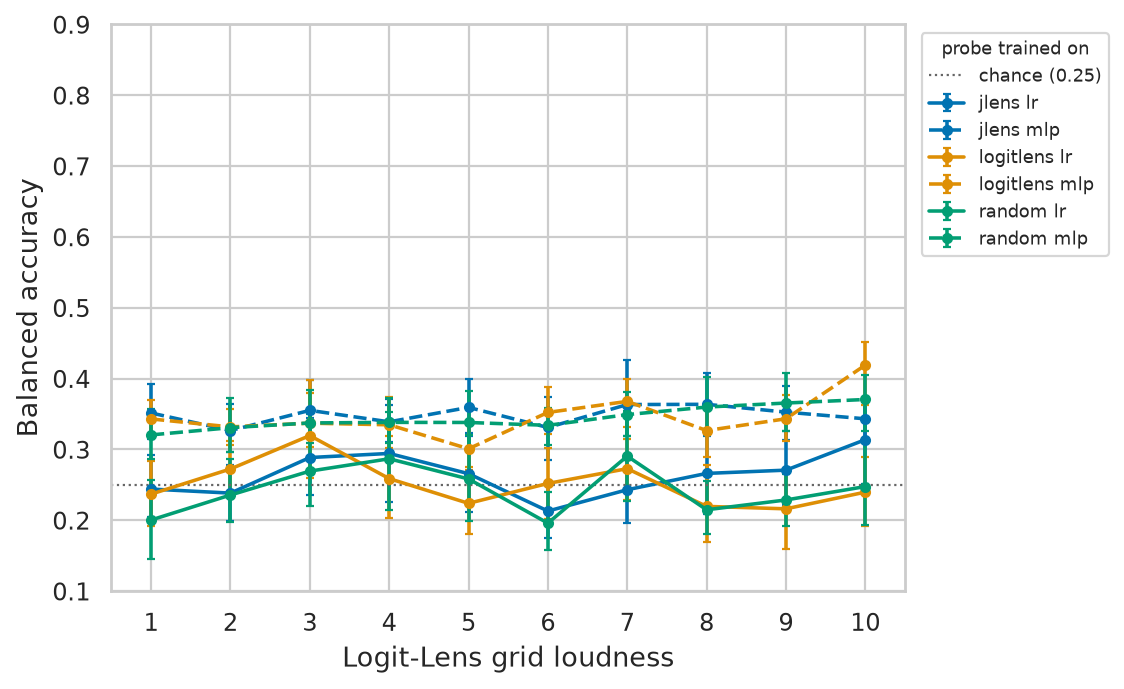} \\
  \end{tabular}
  \caption{Grid probes by grid-loudness decile at $l^*_g$.}
  \label{fig:appx_decile_grid}
\end{figure}

\subsection{Grid-state probes}
\label{appx:grid_results}

Table~\ref{tab:appx_topn_grid} and Figure~\ref{fig:appx_decile_grid} report the grid probes. Linear probes stayed close to chance on every subset and MLP probes remained well below direction-probe accuracy. For Qwen3.6-35B-A3B, J-Lens-selected positions were marginally the most decodable, and no lens showed a trend across deciles. For GPT-OSS-20B, Logit-Lens-selected positions were the most decodable subset for all six probes, and MLP accuracy rose with Logit-Lens but not J-Lens grid loudness, including after grid words were excluded. Because the probes decode so little of the grid state, we do not interpret these differences as evidence for or against either lens (Section~\ref{sec:discussion}). Grid probes binned by direction loudness showed no trend (Figure~\ref{fig:appx_decile_opposite}, bottom).

\subsection{Selected Tokens}
\label{appx:selected_tokens}
As expected, the top tokens in terms of direction J-Lens loudness are the exact direction tokens \texttt{UP}, \texttt{DOWN}, \texttt{LEFT} and \texttt{RIGHT}. Filtering out all $V_d$ tokens from all selections, as well as their $\pm$2 surrounding neighbours, shows us that selected tokens are not trivial selections (Table~\ref{tab:appx_selected_tokens_qwen}).  Most of the loudest remaining tokens are coordinates, brackets and arrows written while the model traces a path through the grid, e.g. \texttt{(1,7)->(2,7)}, where consecutive coordinates implicitly specify a move, or simply where a model is recalling the agent's and the goal's location in the grid. This implies that loudness can find token positions where direction information is present, but not necessarily explicitly verbalized.

\begin{table}[h]
  \centering
  \caption{GPT-OSS-20B, direction, $l^*_d=15$: the $20$ loudest distinct tokens remaining after removing $V_d$ tokens and their $\pm2$ neighbours. $n$: occurrences; context: loudest occurrence, with the token in bold; \textvisiblespace: leading space.}
  \label{tab:appx_selected_tokens}
  \footnotesize
  \begin{tabular}{rlrlr}
    \toprule
    \# & Token & $n$ & Loudest occurrence & Loudness \\
    \midrule
    1 & \texttt{,} & 1{,}051 & \texttt{From~(7~\textbf{,}~1).~We} & $-0.118$ \\
    2 & \texttt{6} & 60 & \texttt{4,8)\#,~(~\textbf{6}~,8)} & $-0.142$ \\
    3 & \texttt{\textvisiblespace{}vertically} & 3 & \texttt{They~are~adjacent~\textbf{vertically}~:~G~below} & $-0.206$ \\
    4 & \texttt{\textvisiblespace{}straight} & 7 & \texttt{blocked.~So~cannot~go~\textbf{straight}~.~Need~det} & $-0.267$ \\
    5 & \texttt{5} & 72 & \texttt{A~down~to~(2,~\textbf{5}~).~Then~left} & $-0.308$ \\
    6 & \texttt{\textvisiblespace} & 31 & \texttt{\_,6\#~Row3:~\textbf{\textvisiblespace}~0\#,1} & $-0.328$ \\
    7 & \texttt{\textvisiblespace{}Lik} & 22 & \texttt{Need~go~around.~\textbf{Lik}~ely~go~down} & $-0.337$ \\
    8 & \texttt{(} & 66 & \texttt{4)~1,~-\textgreater{}~\textbf{(}~2,4} & $-0.349$ \\
    9 & \texttt{\textvisiblespace{}go} & 42 & \texttt{,4)~\#.~So~\textbf{go}~to~(1} & $-0.351$ \\
    10 & \texttt{4} & 62 & \texttt{7\_,8\#.~So~(~\textbf{4}~,7)} & $-0.356$ \\
    11 & \texttt{\textvisiblespace{}horizontally} & 4 & \texttt{They~are~adjacent~\textbf{horizontally}~.~So~move} & $-0.356$ \\
    12 & \texttt{\textvisiblespace{}(} & 267 & \texttt{(6,8)\_,~\textbf{(}~5,7} & $-0.409$ \\
    13 & \texttt{)-\textgreater{}} & 101 & \texttt{left:~(3,2~\textbf{)-\textgreater{}}~(3,} & $-0.433$ \\
    14 & \texttt{2} & 147 & \texttt{(1,7)-\textgreater{}(~\textbf{2}~,7)-\textgreater{}} & $-0.463$ \\
    15 & \texttt{8} & 43 & \texttt{(9,3)-\textgreater{}(~\textbf{8}~,3)-\textgreater{}} & $-0.525$ \\
    16 & \texttt{3} & 219 & \texttt{So~path:~from~(~\textbf{3}~,2)} & $-0.549$ \\
    17 & \texttt{,(} & 2 & \texttt{,(1,3)2~\textbf{,(}~2,3} & $-0.579$ \\
    18 & \texttt{\textvisiblespace{}moving} & 1 & \texttt{Let\textquotesingle{}s~consider~\textbf{moving}~to~(3} & $-0.594$ \\
    19 & \texttt{7} & 90 & \texttt{From~(~\textbf{7}~,6)} & $-0.603$ \\
    20 & \texttt{\textvisiblespace{}det} & 11 & \texttt{So~need~\textbf{det}~our.~From} & $-0.659$ \\
    \bottomrule
  \end{tabular}
\end{table}

\begin{table}[h]
  \centering
  \caption{Qwen3.6-35B-A3B, direction, $l^*_d=27$: the $20$ loudest distinct tokens remaining after removing $V_d$ tokens and their $\pm2$ neighbours. Columns as in Table~\ref{tab:appx_selected_tokens}.}
  \label{tab:appx_selected_tokens_qwen}
  \setlength{\tabcolsep}{4pt}
  \footnotesize
  \begin{tabular}{rlrlr}
    \toprule
    \# & Token & $n$ & Loudest occurrence & Loudness \\
    \midrule
    1 & \texttt{\textvisiblespace} & 164 & \texttt{2:~\_~Col~\textbf{\textvisiblespace}~3:~\_} & $-0.028$ \\
    2 & \texttt{\textvisiblespace{}(} & 68 & \texttt{-~From~\textbf{(}~1,} & $-0.039$ \\
    3 & \texttt{:} & 44 & \texttt{,~1)~Neighbors~\textbf{:}~-~Up} & $-0.052$ \\
    4 & \texttt{\textvisiblespace{}-} & 8 & \texttt{)~is~Right.~\textbf{-}~(1,} & $-0.054$ \\
    5 & \texttt{,} & 167 & \texttt{then~Right~to~(4~\textbf{,}~5)} & $-0.058$ \\
    6 & \texttt{\textvisiblespace{}to} & 13 & \texttt{(1,~3)~\textbf{to}~(2,} & $-0.105$ \\
    7 & \texttt{.} & 52 & \texttt{moves:{*}{*}~1~\textbf{.}~Start~at~(} & $-0.164$ \\
    8 & \texttt{\textvisiblespace{}a} & 3 & \texttt{Is~there~\textbf{a}~path~going~Left} & $-0.165$ \\
    9 & \texttt{(} & 20 & \texttt{5)~is~\_.~\textbf{(}~4,6} & $-0.165$ \\
    10 & \texttt{\textvisiblespace{}-\textgreater{}} & 31 & \texttt{(1,~3)~\textbf{-\textgreater{}}~(2,} & $-0.191$ \\
    11 & \texttt{Moving} & 1 & \texttt{(2,3).~\textbf{Moving}~from~(3} & $-0.208$ \\
    12 & \texttt{\textvisiblespace{}requires} & 1 & \texttt{from~1,3)~\textbf{requires}~ending~with~DOWN} & $-0.249$ \\
    13 & \texttt{1} & 64 & \texttt{2~is~\_)~(~\textbf{1}~,1)} & $-0.265$ \\
    14 & \texttt{\textasciigrave{}} & 6 & \texttt{-~Index~8:~{*}{*}\textbf{\textvisiblespace}{*}{*}~\#} & $-0.283$ \\
    15 & \texttt{Let} & 2 & \texttt{Correct.~\textbf{Let}~\textquotesingle{}s~do~a} & $-0.296$ \\
    16 & \texttt{),} & 11 & \texttt{From~(1,3~\textbf{),}~can~go~DOWN} & $-0.341$ \\
    17 & \texttt{\textvisiblespace{}further} & 1 & \texttt{,~3)~but~not~\textbf{further}~.~At} & $-0.361$ \\
    18 & \texttt{3} & 51 & \texttt{1):~Open~(\_)~(~\textbf{3}~,3):} & $-0.396$ \\
    19 & \texttt{\textvisiblespace{}there} & 1 & \texttt{Blocked.~Is~\textbf{there}~a~path~going} & $-0.422$ \\
    20 & \texttt{)} & 57 & \texttt{-~(1,~3~\textbf{)}~to~(2} & $-0.436$ \\
    \bottomrule
  \end{tabular}
\end{table}

\FloatBarrier
\subsection{Layer sweeps}
\label{appx:layer_sweep}
\begin{figure}[h]
  \centering
  \includegraphics[width=\linewidth]{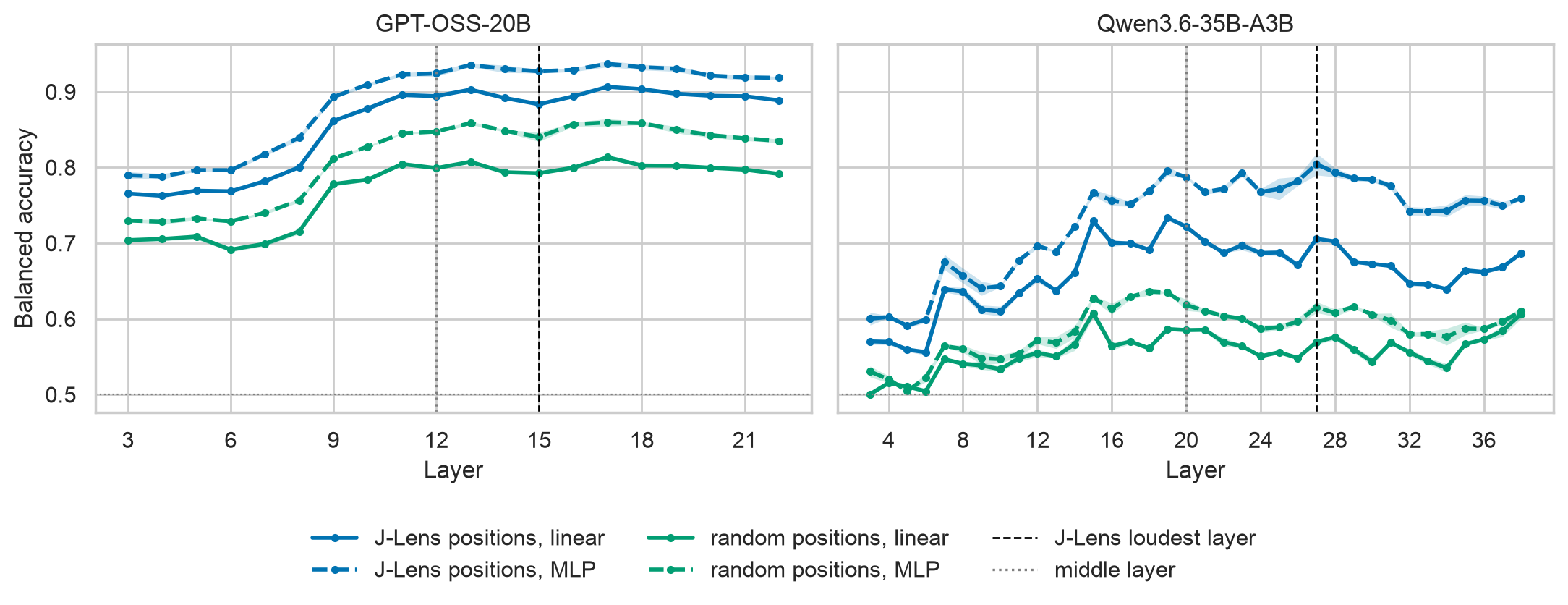}
  \caption{Goal-side probes at every layer, on top-$N$ J-Lens and random positions. Conventions as in Figure~\ref{fig:layer_sweep}; chance is $0.5$.}
  \label{fig:appx_goal_sweep}
\end{figure}

Tables~\ref{tab:appx_sweep_gptoss} and~\ref{tab:appx_sweep_qwen} report the probes of Figures~\ref{fig:layer_sweep} and~\ref{fig:appx_goal_sweep} at every layer. Probes are trained and evaluated on the same positions at every layer: the top-$N$ J-Lens positions, ranked once at $l^*_d$, or random positions. Seeds vary only the initialisation and batch order; the split is fixed.

\begin{table}[h]
  \centering
  \caption{GPT-OSS-20B: per-layer balanced accuracy (final epoch, mean over 5 seeds; max s.d.\ $0.008$; $720$ held-out chains, $628$ for goal side). Columns: training and evaluation positions. \textbf{Bold}: best layer per column; $\star$: J-Lens loudest layer.}
  \label{tab:appx_sweep_gptoss}
  \scriptsize
  \setlength{\tabcolsep}{3.5pt}
  \begin{tabular}{rcccccccc}
    \toprule
    & \multicolumn{4}{c}{Direction (chance $.25$)} & \multicolumn{4}{c}{Goal side (chance $.5$)} \\
    \cmidrule(lr){2-5}\cmidrule(lr){6-9}
    & \multicolumn{2}{c}{J-Lens} & \multicolumn{2}{c}{Random} & \multicolumn{2}{c}{J-Lens} & \multicolumn{2}{c}{Random} \\
    \cmidrule(lr){2-3}\cmidrule(lr){4-5}\cmidrule(lr){6-7}\cmidrule(lr){8-9}
    Layer & Lin. & MLP & Lin. & MLP & Lin. & MLP & Lin. & MLP \\
    \midrule
    3 & .696 & .730 & .578 & .601 & .766 & .790 & .704 & .730 \\
    4 & .697 & .735 & .577 & .608 & .763 & .788 & .706 & .728 \\
    5 & .696 & .741 & .577 & .611 & .769 & .797 & .709 & .733 \\
    6 & .693 & .745 & .569 & .612 & .769 & .796 & .692 & .729 \\
    7 & .701 & .754 & .568 & .615 & .782 & .818 & .699 & .740 \\
    8 & .718 & .772 & .585 & .635 & .800 & .840 & .716 & .756 \\
    9 & .755 & .808 & .619 & .674 & .862 & .893 & .778 & .812 \\
    10 & .775 & .826 & .628 & .692 & .878 & .910 & .784 & .828 \\
    11 & .795 & .842 & .654 & .713 & .896 & .923 & .805 & .845 \\
    12 & .798 & .846 & .658 & .716 & .894 & .924 & .799 & .847 \\
    13 & .805 & .853 & .663 & .719 & .903 & .935 & .807 & .859 \\
    14 & .801 & .852 & .653 & .716 & .892 & .930 & .794 & .849 \\
    15$^\star$ & .800 & .857 & .656 & .716 & .884 & .927 & .792 & .840 \\
    16 & .808 & .867 & .664 & \textbf{.730} & .894 & .929 & .800 & .857 \\
    17 & \textbf{.818} & \textbf{.873} & \textbf{.668} & .725 & \textbf{.907} & \textbf{.937} & \textbf{.814} & \textbf{.860} \\
    18 & .811 & .867 & .660 & .720 & .903 & .932 & .803 & .859 \\
    19 & .806 & .860 & .652 & .706 & .897 & .930 & .802 & .850 \\
    20 & .800 & .856 & .643 & .695 & .895 & .921 & .800 & .843 \\
    21 & .797 & .852 & .643 & .690 & .894 & .919 & .797 & .839 \\
    22 & .795 & .848 & .640 & .685 & .889 & .918 & .792 & .835 \\
    \bottomrule
  \end{tabular}
\end{table}

\begin{table}[h]
  \centering
  \caption{Qwen3.6-35B-A3B: per-layer balanced accuracy (final epoch, mean over 3 seeds; max s.d.\ $0.014$; $52$ held-out chains, $40$ for goal side). Columns: training and evaluation positions. \textbf{Bold}: best layer per column; $\star$: J-Lens loudest layer.}
  \label{tab:appx_sweep_qwen}
  \scriptsize
  \setlength{\tabcolsep}{3.5pt}
  \begin{tabular}{rcccccccc}
    \toprule
    & \multicolumn{4}{c}{Direction (chance $.25$)} & \multicolumn{4}{c}{Goal side (chance $.5$)} \\
    \cmidrule(lr){2-5}\cmidrule(lr){6-9}
    & \multicolumn{2}{c}{J-Lens} & \multicolumn{2}{c}{Random} & \multicolumn{2}{c}{J-Lens} & \multicolumn{2}{c}{Random} \\
    \cmidrule(lr){2-3}\cmidrule(lr){4-5}\cmidrule(lr){6-7}\cmidrule(lr){8-9}
    Layer & Lin. & MLP & Lin. & MLP & Lin. & MLP & Lin. & MLP \\
    \midrule
    3 & .372 & .394 & .319 & .318 & .570 & .600 & .501 & .530 \\
    4 & .356 & .389 & .310 & .311 & .570 & .603 & .515 & .520 \\
    5 & .347 & .381 & .299 & .319 & .560 & .591 & .511 & .505 \\
    6 & .354 & .387 & .296 & .310 & .556 & .599 & .505 & .523 \\
    7 & .413 & .444 & .319 & .327 & .639 & .675 & .547 & .564 \\
    8 & .416 & .448 & .325 & .324 & .636 & .657 & .541 & .560 \\
    9 & .401 & .438 & .311 & .313 & .613 & .640 & .538 & .548 \\
    10 & .396 & .444 & .316 & .316 & .610 & .643 & .534 & .547 \\
    11 & .424 & .470 & .332 & .328 & .634 & .677 & .548 & .554 \\
    12 & .449 & .490 & .353 & .357 & .654 & .697 & .555 & .572 \\
    13 & .435 & .498 & .326 & .340 & .637 & .689 & .551 & .569 \\
    14 & .448 & .536 & .320 & .360 & .661 & .722 & .567 & .583 \\
    15 & .486 & .565 & .349 & .385 & .729 & .767 & \textbf{.608} & .628 \\
    16 & .463 & .558 & .340 & .373 & .701 & .756 & .564 & .614 \\
    17 & .480 & .561 & .337 & .373 & .700 & .752 & .570 & .629 \\
    18 & .478 & .590 & .333 & .376 & .691 & .769 & .561 & \textbf{.636} \\
    19 & .506 & .610 & .351 & .390 & \textbf{.734} & .795 & .587 & .635 \\
    20 & .499 & .595 & .344 & .385 & .722 & .787 & .585 & .619 \\
    21 & .482 & .586 & .345 & .385 & .702 & .768 & .586 & .611 \\
    22 & .470 & .579 & .332 & .387 & .688 & .772 & .570 & .604 \\
    23 & .469 & .596 & .338 & .380 & .697 & .793 & .564 & .601 \\
    24 & .496 & .592 & .345 & .361 & .687 & .768 & .551 & .587 \\
    25 & .478 & .581 & .334 & .356 & .688 & .772 & .556 & .589 \\
    26 & .473 & .602 & .325 & .374 & .671 & .782 & .548 & .597 \\
    27$^\star$ & \textbf{.506} & \textbf{.626} & .345 & \textbf{.395} & .706 & \textbf{.804} & .570 & .616 \\
    28 & .493 & .615 & .339 & .386 & .702 & .794 & .576 & .608 \\
    29 & .486 & .604 & .321 & .370 & .675 & .786 & .560 & .616 \\
    30 & .484 & .603 & .319 & .376 & .673 & .784 & .544 & .606 \\
    31 & .492 & .594 & .313 & .347 & .670 & .776 & .569 & .598 \\
    32 & .477 & .574 & .322 & .341 & .647 & .743 & .556 & .580 \\
    33 & .460 & .555 & .322 & .340 & .646 & .742 & .544 & .580 \\
    34 & .456 & .555 & .321 & .339 & .639 & .743 & .535 & .577 \\
    35 & .475 & .554 & .344 & .352 & .664 & .756 & .567 & .587 \\
    36 & .481 & .539 & .352 & .353 & .662 & .756 & .573 & .587 \\
    37 & .489 & .545 & .358 & .358 & .668 & .750 & .584 & .597 \\
    38 & .499 & .542 & \textbf{.365} & .376 & .687 & .760 & .606 & .610 \\
    \bottomrule
  \end{tabular}
\end{table}

\FloatBarrier
\subsection{The J-Lens as a classifier}
\label{appx:lens_classifier}
\begin{figure}[h]
  \centering
  \includegraphics[width=\linewidth]{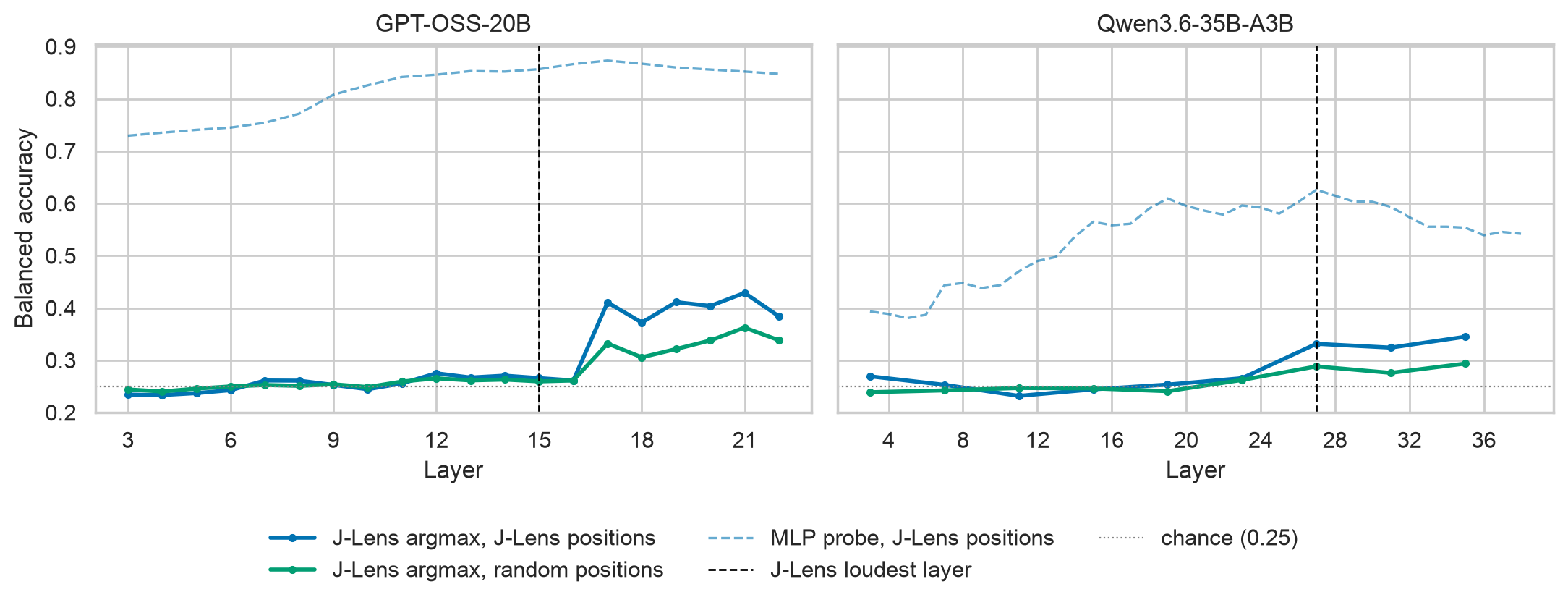}
  \caption{The J-Lens as a direction classifier (argmax over the four action tokens) on J-Lens and random positions, with the MLP probe on J-Lens positions for reference. For Qwen3.6-35B-A3B the lens is fitted only at layers $3, 7, \dots, 35$.}
  \label{fig:appx_lens_classifier}
\end{figure}

As a label-free baseline, we predict the action whose single token (\texttt{LEFT}, \texttt{UP}, \texttt{RIGHT}, \texttt{DOWN}) receives the highest J-Lens log-probability, on the same held-out positions and labels as the direction probes (Figure~\ref{fig:appx_lens_classifier}). At the loudest layers this classifier is close to chance ($.266$ and $.332$): in GPT-OSS-20B its predictions are dominated by a per-token prior (\texttt{DOWN} for $73$--$88\%$ of positions at layers $11$--$16$). It rises above chance only in late layers ($.43$ at most for GPT-OSS-20B, $.35$ for Qwen3.6-35B-A3B), well below trained probes. High direction loudness therefore marks positions where direction information is present, not positions where the lens itself ranks the correct action first.

\FloatBarrier
\subsection{Vocabulary ablation}
\label{appx:vocab_ablation}
\begin{figure}[h]
  \centering
  \includegraphics[width=0.7\linewidth]{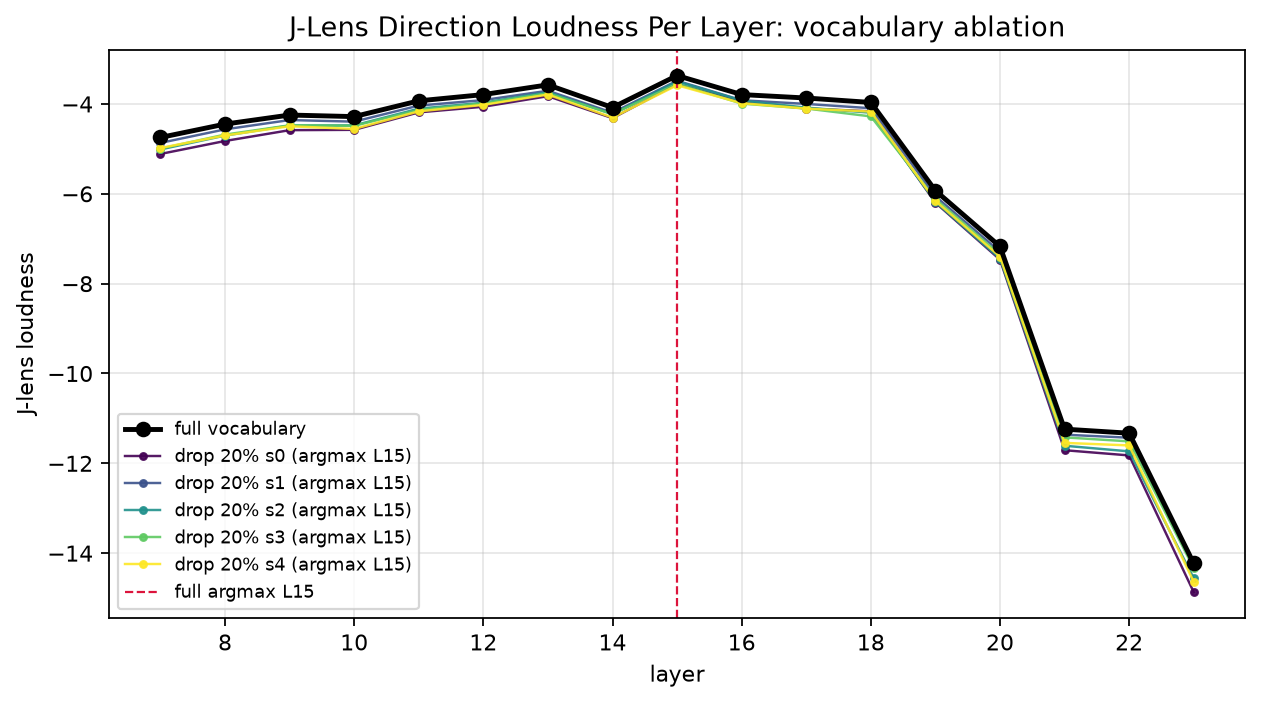}
  \caption{GPT-OSS-20B J-Lens direction loudness per layer for the full vocabulary and five vocabularies with $20\%$ of the words removed.}
  \label{fig:appx_vocab_ablation}
\end{figure}

We removed a random $20\%$ of the pooled direction vocabulary of GPT-OSS-20B ($446 \rightarrow 357$ words; five seeds) and recomputed the J-Lens layer profile on the training chains (Figure~\ref{fig:appx_vocab_ablation}). Removing words shifts the profile down, since less mass is summed, but leaves its shape unchanged: the argmax is layer $15$ and the runner-up layer $13$ for every set, the margin of layer $15$ over layer $13$ is $0.21$--$0.27$ nats against $0.21$ for the full vocabulary, and every perturbed profile correlates with the full one at $r \geq 0.9999$.

\FloatBarrier
\subsection{Goal-side probes by loudness decile}
\begin{figure}[h]
  \centering
  \begin{tabular}{@{}c@{\hspace{2pt}}cc@{}}
    & \small J-Lens & \small Logit Lens \\
    \rotatebox{90}{\small\hspace{1.4em}Direction} &
    \includegraphics[width=0.42\linewidth]{figures/loudness_probe_performance_analysis/left-right-direction-gptoss-probes/gptoss_goal-side-probes_jlens_direction-loudness.png} &
    \includegraphics[width=0.42\linewidth]{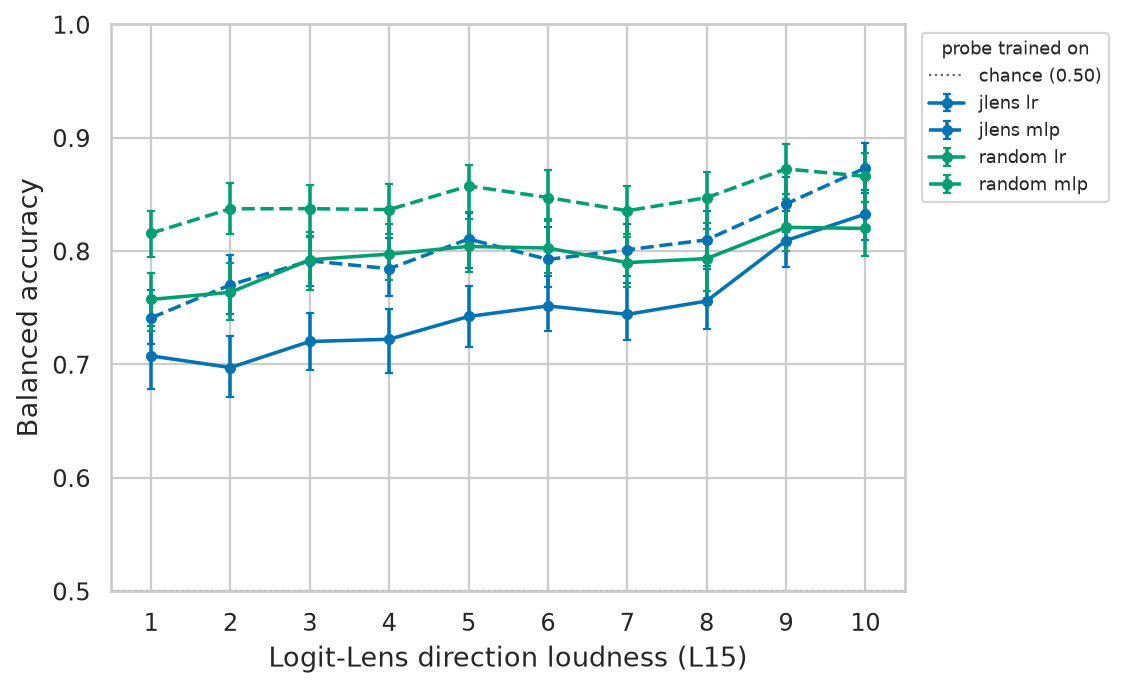} \\
    \rotatebox{90}{\small\hspace{0.2em}Direction, no words} &
    \includegraphics[width=0.42\linewidth]{figures/loudness_probe_performance_analysis/left-right-direction-gptoss-probes/gptoss_goal-side-probes_jlens_direction-loudness_no-direction-words-r2.png} &
    \includegraphics[width=0.42\linewidth]{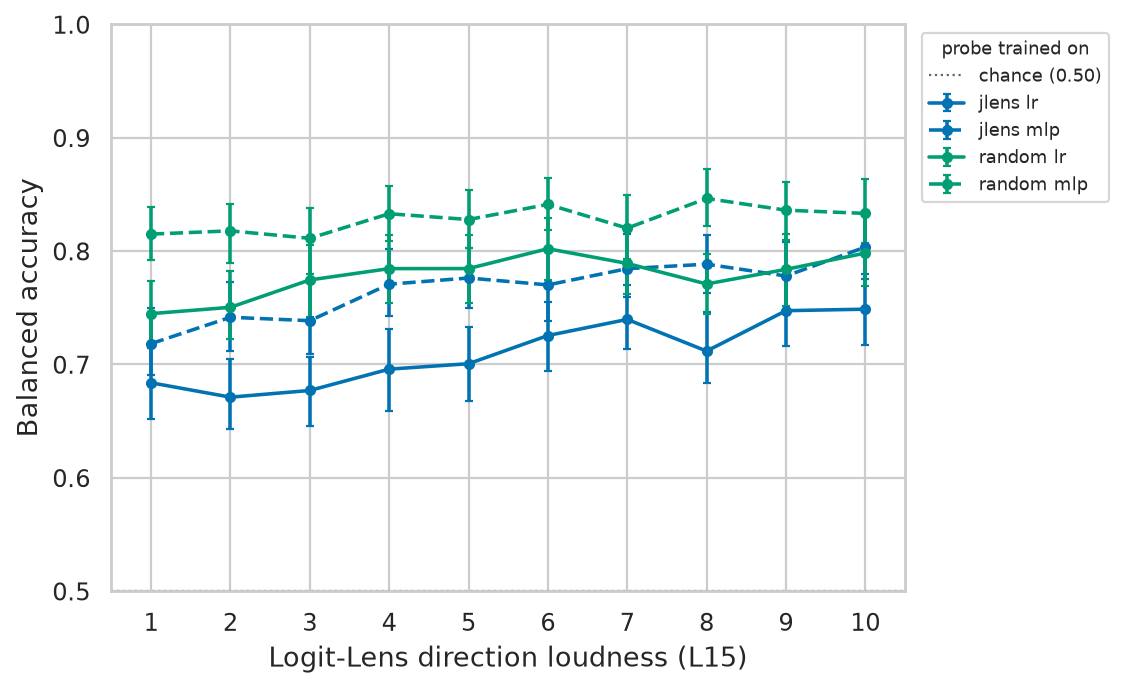} \\
    \rotatebox{90}{\small\hspace{2.4em}Grid} &
    \includegraphics[width=0.42\linewidth]{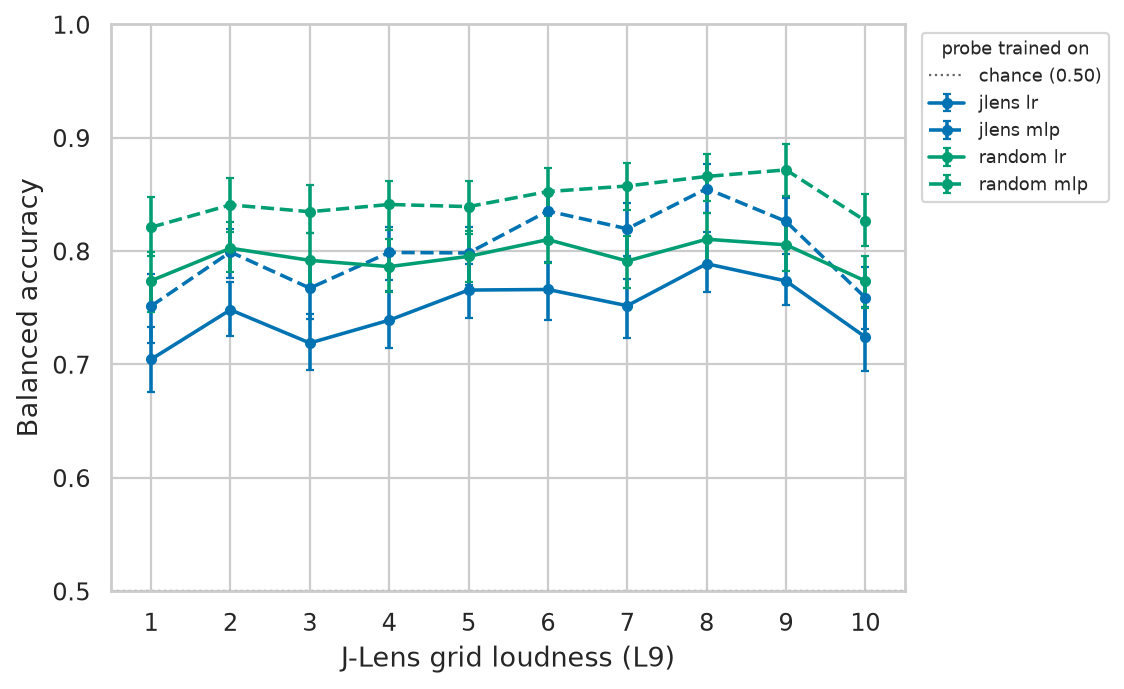} &
    \includegraphics[width=0.42\linewidth]{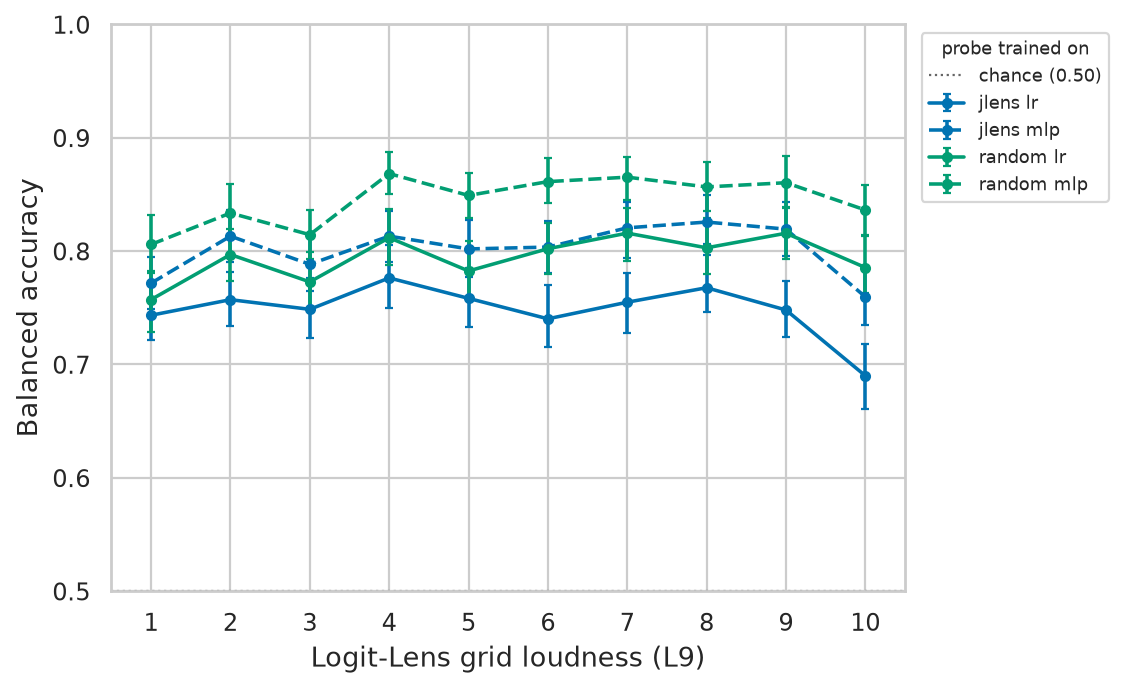} \\
  \end{tabular}
  \caption{GPT-OSS-20B goal-side probes by loudness decile. Rows: direction loudness, direction loudness with direction words and their $\pm2$ neighbours removed, grid loudness. Columns: lens. Chance is $0.5$.}
  \label{fig:appx_goal_deciles}
\end{figure}

Figure~\ref{fig:appx_goal_deciles} bins the GPT-OSS-20B goal-side probes (layer $15$, seed $42$) by direction and grid loudness, on held-out random positions ($12{,}553$ tokens, $628$ chains). Accuracy rises with J-Lens direction loudness, also after removing direction words, more weakly with Logit-Lens direction loudness, and not at all with grid loudness.

\FloatBarrier
\end{document}